\documentclass[journal,twoside,web]{ieeecolor}
\usepackage{tmi}
\usepackage{cite}
\usepackage{amsmath,amssymb,amsfonts}
\usepackage{algorithmic}
\usepackage{graphicx}
\usepackage{textcomp}
\usepackage{subfigure}

\DeclareMathOperator{\realset}{\mathbb{R}}
\DeclareMathOperator{\intset}{\mathbb{Z}}

\usepackage{booktabs}
\usepackage{bm}
\usepackage{mathtools}
\usepackage{dsfont}
\usepackage{algorithm}
\usepackage{algorithmic}
\usepackage{multirow}
\let\labelindent\relax
\usepackage{enumitem}
\usepackage{xcolor}
\usepackage{pifont}
\newcommand{\cmark}{\ding{51}}%
\newcommand{\xmark}{\ding{55}}%

\usepackage[capitalize]{cleveref}
\crefname{section}{Sec.}{Secs.}
\Crefname{section}{Section}{Sections}
\Crefname{table}{Table}{Tables}
\crefname{table}{Table.}{Tables.}

\crefformat{equation}{#2(#1)#3}
\Crefformat{equation}{#2(#1)#3}
\crefmultiformat{equation}{#2(#1)#3}{ and #2(#2)#3}{, #2(#2)#3}{ and #2(#2)#3}
\Crefmultiformat{equation}{#2(#1)#3}{ and #2(#2)#3}{, #2(#2)#3}{ and #2(#2)#3}
\crefrangeformat{equation}{(#3#1#4) to (#5#2#6)}

\DeclarePairedDelimiterX{\infdivx}[2]{(}{)}{%
  #1\;\delimsize\|\;#2%
}

\newcommand{\infdiv}{D_{\mathrm{KL}}\infdivx}
\newcommand{\mb}[1]{\mathbf{#1}}

\newcommand{\Expect}[2]{\mathop{\mathbb{E}} \limits_{#1}\left[ #2 \right]}

\def\BibTeX{{\rm B\kern-.05em{\sc i\kern-.025em b}\kern-.08em
    T\kern-.1667em\lower.7ex\hbox{E}\kern-.125emX}}
\begin{document}

\title{Improving Robustness and Reliability in Medical Image Classification with Latent-Guided Diffusion and Nested-Ensembles\thanks{© 2025 IEEE. Personal use of this material is permitted. Permission from IEEE must be obtained for all other uses, in any current or future media, including reprinting/republishing this material for advertising or promotional purposes, creating new collective works, for resale or redistribution to servers or lists, or reuse of any copyrighted component of this work in other works.}}
\author{Xing Shen, Hengguan Huang, Brennan Nichyporuk, and Tal Arbel
\thanks{This work was supported in part by the Natural Sciences and Engineering Research Council of Canada, in part by the Canadian Institute for Advanced Research (CIFAR) Artificial Intelligence Chairs Program, in part by the Mila—Quebec Artificial Intelligence Institute Technology Transfer Program, in part by the Mila—Google Research Grant, in part by Calcul Quebec, in part by the Digital Research Alliance of Canada, and in part by the Canada First Research Excellence Fund, awarded to the Healthy Brains, Healthy Lives initiative at McGill University.}
\thanks{Xing Shen is at the Centre for Intelligent Machines, McGill University, Montreal, QC H3A 0G4 Canada, and a student at Mila -- Quebec AI Institute, Montreal, QC H2S 3H1 Canada (e-mail:xing.shen@mail.mcgill.ca). }
\thanks{Hengguan Huang is an Assistant Professor at Department of Public Health, Section for Health Data Science and AI, University of Copenhagen, 1172 København, Denmark. Corresponding author (e-mail:hengguan.huang@sund.ku.dk).}
\thanks{Brennan Nichyporuk is a Research Scientist at Mila -- Quebec AI Institute, Montreal, QC H2S 3H1 Canada. He is also an affiliate member of  McGill University, Montreal, QC H3A 0G4 Canada (e-mail:nichypob@mila.quebec).}
\thanks{Tal Arbel is a Professor at McGill University, and a member of the Centre for Intelligent Machines, McGill University, Montreal, QC H3A 0G4 Canada. She is also a Fellow of the Canadian Academy of Engineering, a CIFAR AI Chair, and Core Member of Mila -- Quebec AI Institute, Montreal, QC H2S 3H1 Canada. Corresponding author (e-mail: tal.arbel@mcgill.ca).}}

\maketitle

\begin{abstract}
 Once deployed, medical image analysis methods are often faced with unexpected image corruptions and noise perturbations. These unknown covariate shifts present significant challenges to deep learning based methods trained on ``clean'' images. This often results in unreliable predictions and poorly calibrated confidence, hence hindering clinical applicability. While recent methods have been developed to address specific issues such as confidence calibration or adversarial robustness, no single framework effectively tackles all these challenges simultaneously. To bridge this gap, we propose \textit{LaDiNE}, a novel ensemble learning method combining the robustness of Vision Transformers with diffusion-based generative models for improved reliability in medical image classification. Specifically, transformer encoder blocks are used as hierarchical feature extractors that learn invariant features from images for each ensemble member, resulting in features that are robust to input perturbations. In addition, diffusion models are used as flexible density estimators to estimate member densities conditioned on the invariant features, leading to improved modeling of complex data distributions while retaining properly calibrated confidence. Extensive experiments on tuberculosis chest X-rays and melanoma skin cancer datasets demonstrate that LaDiNE achieves superior performance compared to a wide range of state-of-the-art methods by simultaneously improving prediction accuracy and confidence calibration under unseen noise, adversarial perturbations, and resolution degradation.
\end{abstract}


\section{Introduction}
\IEEEPARstart{I}{n} the rapidly evolving domain of medical imaging analysis, deep learning has led to enormous advances in many clinical domains of interest~\cite{esteva2017dermatologist, gulshan2016development, ardila2019end, bejnordi2017diagnostic, de2018clinically, litjens2017survey, qiu2016initial, tajbakhsh2016convolutional, milletari2016v}, notably in tasks such as detection of diabetic retinopathy in eye fundus images~\cite{gulshan2016development}, classification of skin cancer~\cite{esteva2017dermatologist}, and identification of cancerous regions in mammograms~\cite{qiu2016initial}. Despite the fact that recent methods have achieved unprecedented success in controlled experimental settings, their fundamental building blocks – deep neural networks (DNNs) – are known to be sensitive to slight distribution changes or covariate shifts, and vulnerable to attacks \cite{hendrycks2018benchmarking,szegedy2014intriguingpropertiesneuralnetworks,navarro2021evaluating}. As a result, their application to real-world clinical contexts often results in significant performance degradation, including inaccurate predictions and poorly calibrated confidence estimates.

Data augmentation is a widely used tool for improving the generalization and robustness of DNNs. In the field of medical imaging, however, where datasets are typically smaller than natural image datasets especially for rare diseases~\cite{roglin2022improving}, conventional data augmentation strategies may not always be suitable and can even lead to degradation in the performance of DNNs~\cite{takase2021self}. Designing suitable augmentation strategies from small medical imaging datasets can be challenging, requiring careful design choices in order to incorporate suitable inductive biases that lead to robust and generalizable DNN architectures and learning algorithms.

Real-world medical imaging applications, such as medical image classification, often face unpredictable and complex covariate shifts that are difficult to anticipate during training~\cite{matta2024systematic}, leading to significant challenges in maintaining reliable performance of DNN-based methods. In the field of machine learning, ensemble methods are popular choices to enhance robustness and generalization, as they combine the predictions of multiple models, effectively reducing variance and mitigating overfitting~\cite{breiman1996bagging}. By leveraging the strengths of diverse models, ensemble techniques such as bagging, boosting, and stacking can achieve better and more robust predictive performance over a single model~\cite{Mienye2022A}. In addition, deep ensemble methods have been shown to improve both predictive performance and the quality of uncertainty estimates by training multiple DNNs independently and averaging their predictions~\cite{lakshminarayanan2017simple}. Other frameworks involve developing algorithms to select better ensemble members and distribute input-dependent weights to each member, aiming to reduce the effect of weak members and providing performance gains~\cite{yang2021two,Pacheco2020Learning}. Despite the potential robustness offered by ensemble methods, their application in medical image classification to address covariate shifts and confidence calibration remains limited. While in natural image classification, existing ensemble methods often rely on simplified component distribution assumptions, such as Gaussian distributions or deterministic mappings to the parameters of a categorical distribution~\cite{lakshminarayanan2017simple}, which may not adequately capture the complexity and heteroscedasticity of real-world medical imaging data.
 
This paper introduces {\it \textbf{LaDiNE}}, \textbf{La}tent-guided \textbf{Di}ffusion \textbf{N}ested-\textbf{E}nsembles, a robust ensemble learning model designed for mitigating the aforementioned challenges in medical image classification. LaDiNE is a parametric mixture model that incorporates transformers and diffusion models, leveraging their recent success in medical imaging contexts~\cite{shamshad2023transformers,kazerouni2023diffusion}. It encodes invariant and informative features as latent variables, and performs functional-form-free inference to estimate the predictive distribution. Specifically, a subset of transformer encoder blocks of a Vision Transformer (ViT) is used as a hierarchical feature extractor that learn invariant features from images for each mixture component. The diffusion models are used as flexible density estimators to estimate the component densities conditioned on the invariant features. In our formulation, each mixture component is interpreted equivalently as a Bayesian network that encodes the dynamics of observed (e.g., images) and latent variables. LaDiNE is specifically designed to be: (i) Robust to covariate shift; (ii) Provide calibrated confidence estimates; (iii) Be resilient to gradient-based adversarial attacks.

Extensive experiments are performed on the Tuberculosis chest X-ray classification benchmark~\cite{rahman2020reliable} and the ISIC skin cancer classification benchmark~\cite{rotemberg2021patient}. We split the original datasets into training, validation, and testing sets. Models are trained on the original training set and then evaluated on a perturbed version of the test set, one with complex simulated unseen covariate shifts. Results indicate that  {\it \textbf{LaDiNE}} performs substantially better than popular baselines in terms of prediction accuracy and prediction confidence calibration, under a variety of challenging covariate shift conditions. Our contribution is threefold:
\begin{itemize}[labelwidth=!, topsep=0pt, leftmargin=0.4cm]
     \item[1.] This paper introduces LaDiNE, a novel ensemble learning method that encodes invariant features and estimates the predictive component distribution without specific assumptions regarding its component functional form using, for the first time, diffusion models. This additional component enables flexible and expressive modeling of complex distributions,  especially important in the with limited medical data.
     \item[2.] Extensive evaluations demonstrate that the proposed method outperforms a wide variety of state-of-the-art (SOTA) methods in terms of classification accuracy under diverse covariate shift scenarios, including (i) images with Gaussian noise injection, (ii) images with lower resolution, (iii) images with lower color contrast, and (iv) images with adversarial perturbation. Empirical evidence shows significant improvements in classification accuracy over baseline methods. Detailed ablation studies further justify the design choices made in the paper.
     \item[3.] Extensive experimentation indicates that the proposed method provides better calibrated predictions than competing methods. Instance-level prediction uncertainties are evaluated under severe perturbations of the input images and are shown to be correct when more certain, as desired.
\end{itemize}

\section{Related Work}
This section summarizes existing work on robustness learning in medical imaging, focusing on methods based on transformers and diffusion models.

\textbf{Transformers in Medical Imaging.}\; While transformer-based models have been less extensively studied in medical image analysis compared to their widespread application in computer vision, they hold significant potential for capturing complex spatial relationships and the variability inherent in clinical data~\cite{shamshad2023transformers}. Some recent works have shown how incorporating transformers helps to improve prediction accuracy in various medical imaging tasks. Peiris \textit{et al}.~\cite{peiris2022robust} proposes a transformer-based architecture that can encode local and global spatial cues for 3D tumor segmentation and exhibits robust performance against the presence of image artifacts. Chen \textit{et al.}~\cite{chen2022gashis} integrates multi-scale feature extraction into the transformer to achieve improved performance in image-based gastric cancer detection, with the ability to be robust to noise. Wang \textit{et al.}~\cite{wang2021dudotrans} uses transformers to capture X-ray sinograms’ global characteristics and achieve enhanced performance against artifacts in sparse-view CT reconstruction. Almalik \textit{et al.} \cite{almalik2022self} extends this direction by leveraging intermediate representations from ViT blocks to improve robustness against adversarial attacks. Although these transformer-based methods demonstrate some degree of robustness to specific forms of noise and artifacts, they do not explicitly address confidence calibration and uncertainty quantification under noisy conditions or distribution shifts in input images. As a result, their ability to assist in informed clinical decision-making under unknown covariate shifts remains limited.

\textbf{Diffusion Models in Medical Imaging.}\; Diffusion models have recently emerged as powerful generative models for medical imaging applications due to their ability to generate high-quality images, while remaining robust to distribution shifts~\cite{kazerouni2023diffusion,dhariwal2021diffusion}. While most of the current work is focused on medical image generation and reconstruction with diffusion models~\cite{pinaya2022brain,moghadam2023morphology,yoon2023sadm,song2021solving,xie2022measurement}, some methods have been developed for medical image segmentation via generating the segmentation mask~\cite{wu2024medsegdiff,wolleb2022diffusion}. In other work, Li \textit{et al.}~\cite{li2023zero} uses frequency-domain filters to guide the diffusion model for structure-preserving image translation, achieving robust generalization capability. Kim \textit{et al.}~\cite{kim2022diffusion} incorporates diffusion models into a representation learning framework for vessel segmentation and shows superior results on noisy data. In the field of machine learning, recent work by Clark and Jaini demonstrates that a text-to-image diffusion model can function as a zero-shot classifier \cite{clark2024text}. Han \textit{et al.} \cite{han2022card} further introduces a diffusion-based conditional generative framework tailored for classification and regression on natural images and tabular data, showing strong performance in uncertainty estimation. Despite their robustness and ability to provide instance-level predictive uncertainty \cite{chendiffusion,han2022card}, which can aid in informed clinical decision-making, their application in medical image classification has yet to be explored.

\section{The Proposed Method: LaDiNE}
\label{sec:method}
This work focuses on establishing a robust and generalizable model for medical image classification, where a model trained on ``clean'' images would be required to be robust to substantial, unseen covariate shifts. The proposed framework consists of several important components: (i) transformer encoders (TEs) derived from one Vision Transformer (ViT)~\cite{dosovitskiyimage}, (ii) conditional diffusion models (CDM), and (iii) feed-forward networks (FFNs). In this section, we first give a high-level overview of the proposed method. Then we describe the notation of variables and the computational paths involved in each neural network. Finally, we introduce the proposed probabilistic model in \cref{subsec:ppm} and the training procedure for those neural networks in \cref{subsec:training}.

\begin{figure*}
  \centering
  \includegraphics[width=1\linewidth]{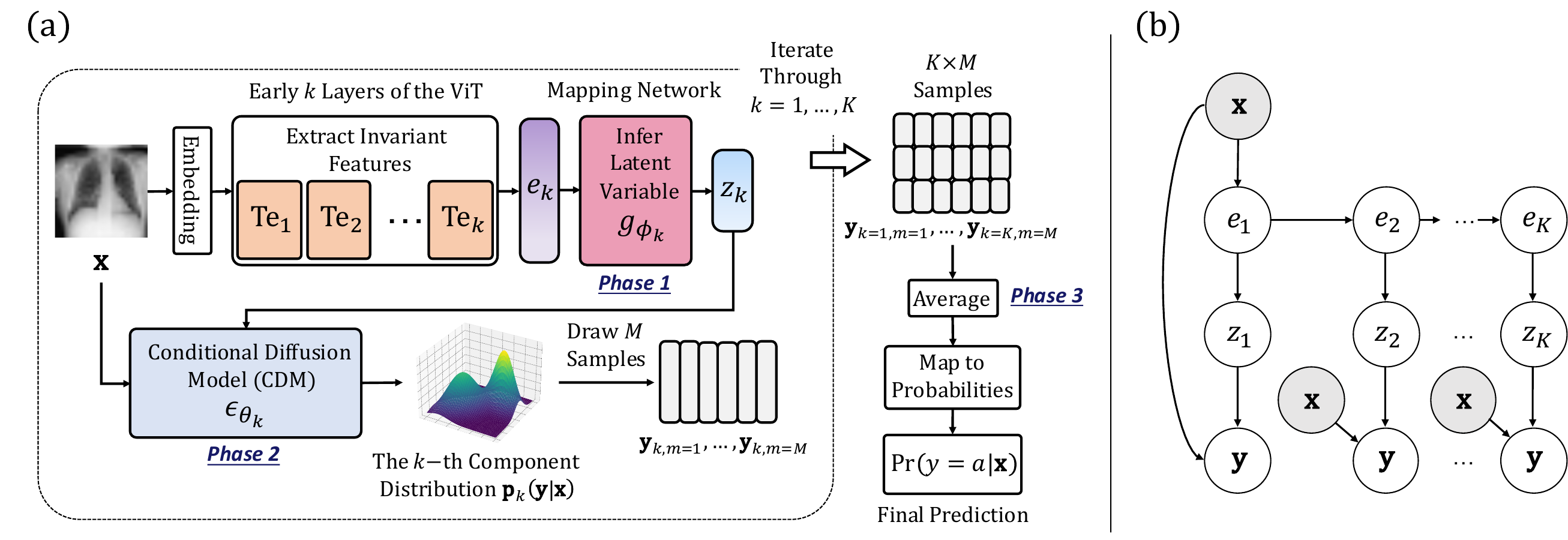}
  \caption{An illustration of the proposed model from two perspectives: (a) The flowchart shows the workflow of the proposed model in three phases. In Phase 1, the transformer encoders and the mapping network $g_{\phi_k}$ compute the latent variable $z$ from the image $\mb x$. In Phase 2, a conditional diffusion model estimates the predictive component distribution. $M$ samples are drawn from this distribution. In Phase 3, $M$ samples are extracted from each of the $K$ ensemble members. These samples are aggregated to form the final prediction. (b) This directed acyclic graph shows the dependency of each variable within the (unrolled) probabilistic model. Here $\mb x$ (observed in grey) and $\mb y$ are the input image and its predicted label, respectively. $e_k$ denote the image embedding and $z_k$ denotes the latent variable in the $k$-th ensemble member. The mixture weight variables are omitted as the components are equally weighted. In this graph, every directed edge shows dependencies. For example, $\mb x \rightarrow e_1$ means that $e_1$ depends on $\mb x$, and the local Markovian property holds.}
  \label{fig:prediction}
\end{figure*}

\textbf{Overview.}\; LaDiNE is an ensemble deep learning model specifically designed to be robust to covariate shifts, while providing high-quality prediction confidence. To this end, LaDiNE (i) leverages early transformer encoders from only one ViT and mapping networks in order to learn image representations that are robust across different environments and (ii) estimates component distributions through conditional diffusion models, free from fixed distributional assumptions (see \cref{fig:prediction} (a) for an illustration of the proposed method).

\textbf{Notations.}\; An image input is denoted  $\mb{x}\in\realset^{H\times W\times C}$, where $(H\times W)$ is the resolution of the image and $C$ is the number of channels. Its corresponding label is denoted as $y$ (class index) and $\mb{y}$ (one-hot encoded vector). We assume that the observational data pair $\langle \mb{x}, y \rangle$ is sampled from the joint distribution $p_\text{train}(\mb{x}, y)$, which serves as the training data for the model. In a covariate shift setting, the test data points $\langle \mb{x}', y' \rangle$ are sampled from a different distribution, denoted as $p_\text{test}(\mb{x}', y')$. The covariate shift assumption implies that the conditional distribution of labels given the input remains unchanged, i.e., $p_\text{train}(y|\mb{x}) = p_\text{test}(y'|\mb{x}')$, but the marginal distribution of the inputs differs, i.e., $p_\text{train}(\mb{x}) \neq p_\text{test}(\mb{x}')$. As a result, we have a divergence measure $D(p_\text{train}(\mb{x}), p_\text{test}(\mb{x}')) \neq 0$.

\textbf{Vision Transformer and Transformer Encoders.}\; We follow the architecture described in \cite{dosovitskiyimage}, and introduce some additional notation used here. The Vision Transformer (ViT) is the stack of $L$ transformer encoder (TE) blocks, where each TE block acts as a deterministic mapping from the input sequence to the output sequence. We define $\text{TE}:\realset^{n\times d}\rightarrow\realset^{n\times d}$ to represent this mapping. When computing TE, the input image $\mathbf{x}\in\realset^{H\times W\times C}$ is first divided into patches. Those patches are then flattened and projected into a lower-dimensional embedding space. This process, including the addition of positional encoding, can be simplified into a single embedding function $e_0=\text{Emb}(\mb x)$. Before feeding into the first TE block, a class token (with its own positional encoding)  is appended to the embedding $e_0$. For simplifying the whole process, we denote another function $\text{Te}(\cdot)$ as:
\begin{gather}
    \text{Te}(e_k):=\text{TE}(\{\text{class token},e_k\}),
\end{gather}
here $e_k$ is the embedding output (excluding the class token) generated by the $(k-1)$-th TE block. The class token generated by the last TE block is used for classification with a feed-forward network (FFN).

\begin{figure}[]
  \centering
  \subfigure{\includegraphics[width=0.48\linewidth]{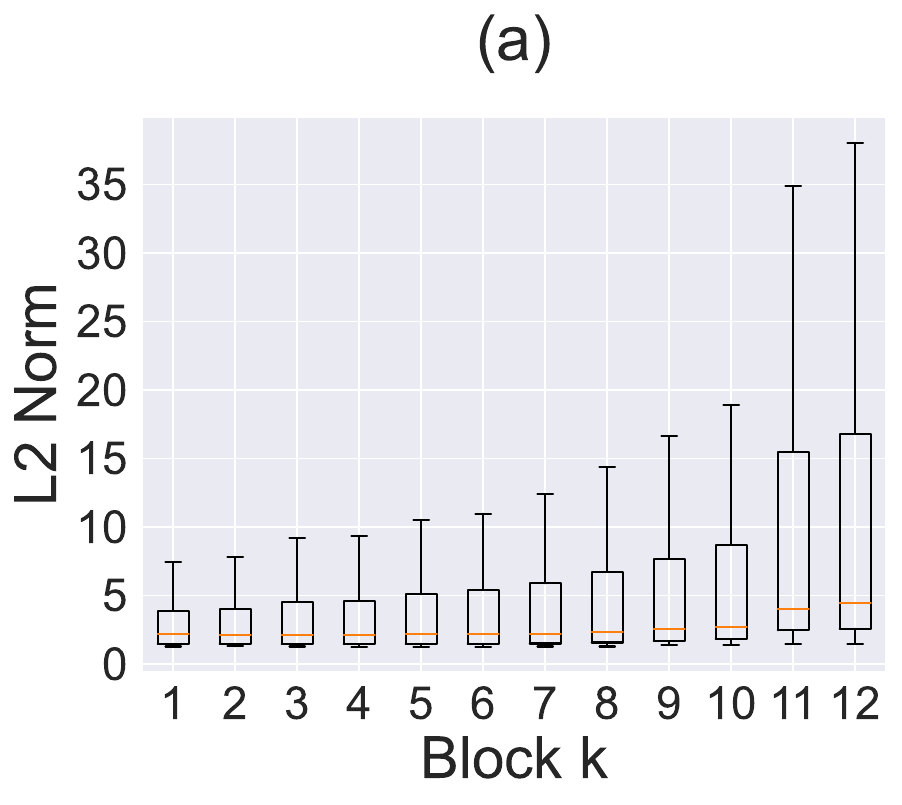}}
  \subfigure{\includegraphics[width=0.48\linewidth]{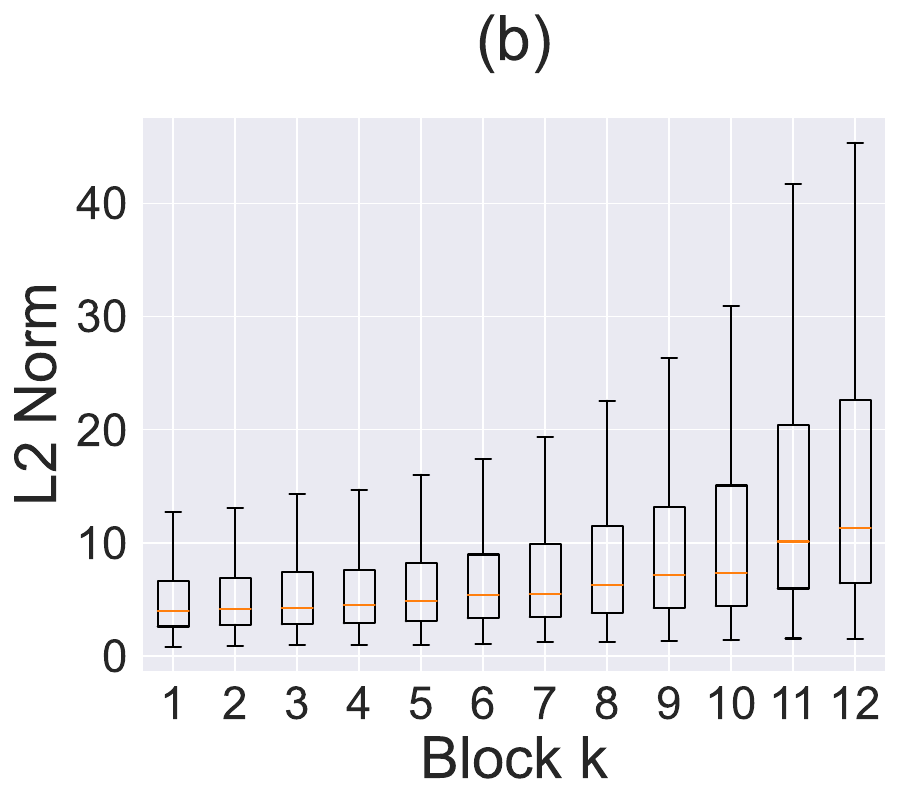}}
  \subfigure{\includegraphics[width=0.48\linewidth]{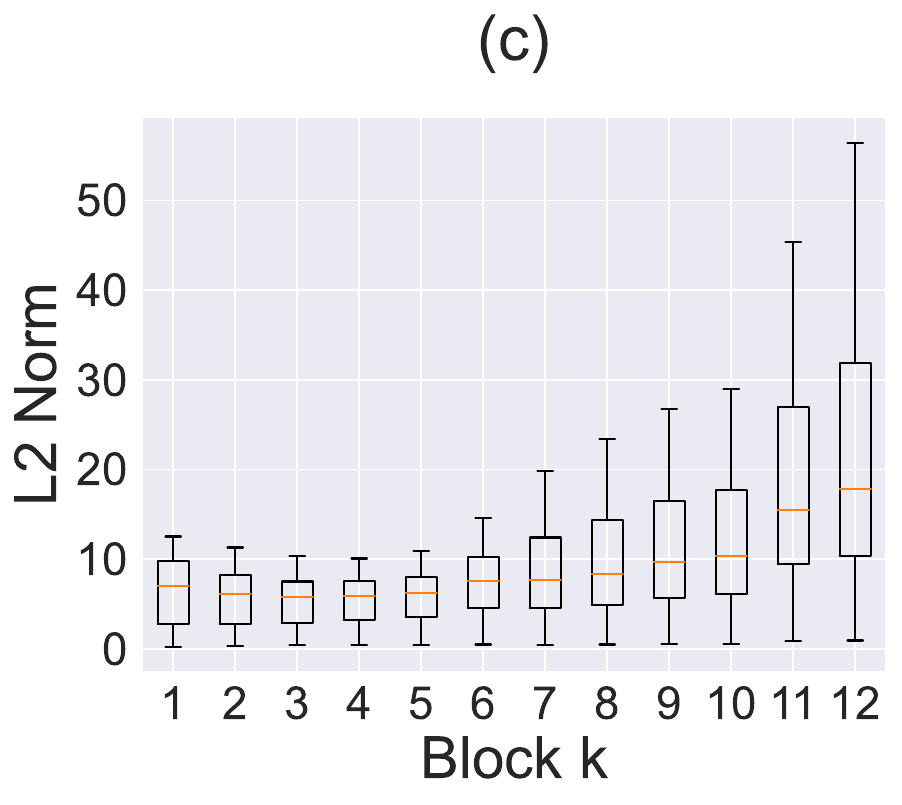}}
  \subfigure{\includegraphics[width=0.48\linewidth]{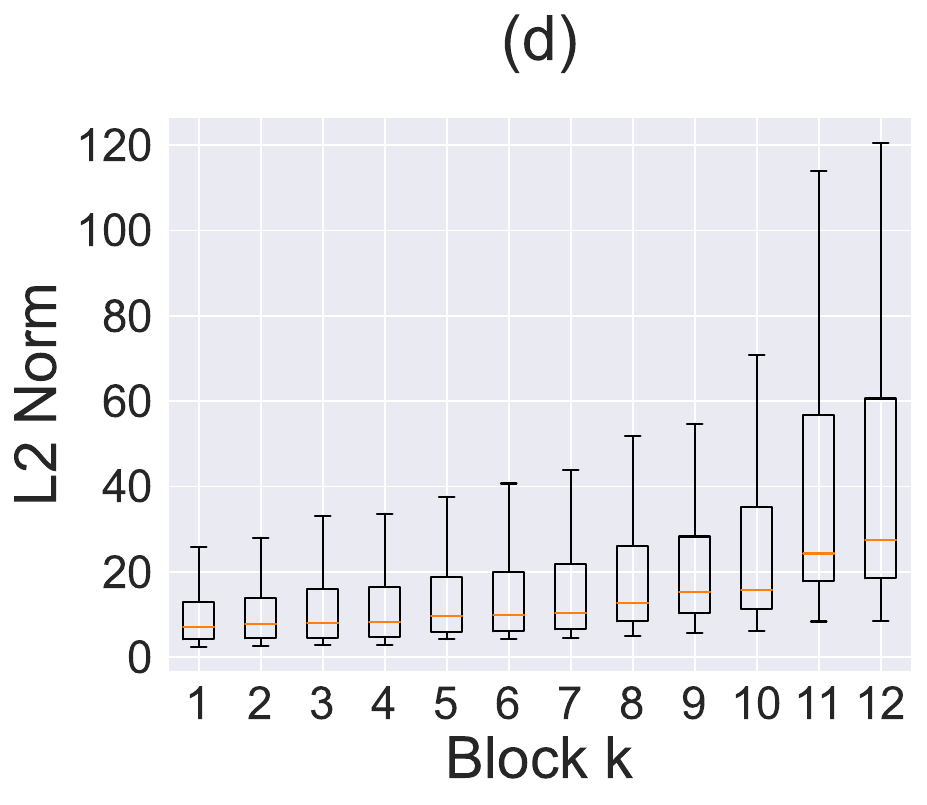}}
  \caption{The Euclidean distance between the token sequence of image variants (under a range of conditions) and its original copy increases as going deeper into the encoder block hierarchy under (a) noisy, (b) lower-resolution, (c) lower-contrast contexts, and (d) adversarial attack.}
  \label{fig:l2}
\end{figure}

In Almalik \textit{et al.} and Walmer \textit{et al.}, the authors presented an investigation of the invariance to adversarial noise \cite{almalik2022self} and informative hidden layers in ViTs \cite{walmer2023teaching}. Extending these findings, we investigate whether the representations extracted from the early TE blocks in ViTs remain robust under a range of image perturbations, beyond adversarial noise considered in prior work \cite{almalik2022self}. To this end, a number of experiments are performed. \cref{fig:l2} depicts the results where the L2 norms of the differences between clean input and noisy input representations are shown, under four types of covariate shifts. The results indicate a clear pattern where, under all conditions, early TE blocks learn more invariant features than the deeper blocks. This finding motivates the use of early TE blocks in the predictions in order to improve robustness performance. Specifically, here $\text{TE}_k$ is included where $k=1,2,\dotsc ,K$ and $K<L$.

\textbf{Mapping to Latents.}\; In addition to using FFNs within the ViT, FFNs are used to map the embedding $e$ (e.g. $e_1$ defined previously)  to a latent $z$ for subsequent computations. As such, it is named as a {\it mapping network}. In order not to confuse this mapping with the FFNs used in the ViT, it is denoted as a function $g:\realset^{N\times d}\rightarrow\realset^{d_\text{latent}}$ as follows:
\begin{gather}
    z=g(e)=\text{FFN}(e). \label{eq:g}
\end{gather}

\textbf{Estimating Distribution with Diffusion Models.}\; In \cite{han2022card}, the diffusion model (DM) models a conditional distribution free of predefined functional-forms with a single covariate. Here a conditional DM (CDM) is defined with several covariates: the latent $z$ and the image input $\mb x$, together with the response $\mb y$. This enables sampling from the probability density function $p(\mb y| z,\mb x)$.

\subsection{Probabilistic Predictive Model}
\label{subsec:ppm}
The proposed method is visually depicted in \cref{fig:prediction} (b), which provides a graphical representation of the dependencies among variables. This representation enables the factorization of the predictive distribution into conditional components. In this section, we describe the parametrization of each factorized conditional distribution.

\textbf{Predictive Distribution.}\; The proposed predictive model is then defined as a mixture model composed of $K$ components,  or {\it ensemble members}:
\begin{gather}
    p(\mb y| \mb x, \Theta)=\sum_{k=1}^K\pi_k\underbrace{\int\dotsb\int p(\mb y,z_k,e_{1:k}|\mb x)\: dz_kde_{1:k}}_{\mb p_k(\mb y|\mb x)}
\end{gather}
where $\Theta$ denotes all parameters in the mixture, $\pi_k$ is the mixture weight for the $k$-th component distribution $\mb p_k(\mb y|\mb x)$ with $\sum_{k=1}^K \pi_k = 1$ and $\pi_k \geq 0$.
The $k$-th predictive component can be further factorized as:
\begin{align}
    \mb p_k(\mb y|\mb x)=&\int\dotsb\int p(\mb y| z_k,\mb x)p(z_k|e_k) \notag \\ 
    &\quad\quad\quad\; \prod_{i=2}^k p(e_i|e_{i-1})p(e_1|\mb x) \:dz_kde_{1:k}.
\end{align}
This factorization allows us to compute the predictive density by breaking it down into a series of conditional distributions. The next step is to parameterize these conditional distributions so that the model can learn from the data.

\textbf{Parameterization.}\; When there is no particular prior belief in the contribution of each mixture component, each component is equally weighted, such that $\pi_k=K^{-1}$. With deterministic functions $\text{Te}(\cdot)$ and $g(\cdot)$, the component distribution becomes:
\begin{align}
    \mb p&_{k}(\mb y|\mb x)=\int\dotsb\int p_{\theta_k}(\mb y| z_k,\mb x)\delta(z_k-g_{\phi_k}(e_k)) \notag \\
    &\prod_{i=2}^k \delta(e_i-\text{Te}_i(e_{i-1})) \delta(e_1-\text{Te}_1(\text{Emb}(\mb x))) \; dz_kde_{1:k},
\end{align}
where $\delta(\cdot)$ is the Dirac Delta function, and $\text{Emb}(\mb x)$ is the embedding step to produce $e_0$. Here, the conditional distribution $p_{\theta_k}(\mb y| z_k,\mb x)$ modeled by the DM is parameterized by $\theta_k$, and the mapping network $g_{\phi_k}(\cdot)$ is parameterized by $\phi_k$. The notation of the parameters in $\text{Emb}(\cdot)$ and $\text{Te}(\cdot)$ are omitted here as they are packed in the TE blocks' parameters, which are estimated along the training of the ViT. After simplification of the $\delta$ distribution, the model becomes:
\begin{align}
    \mb p&_{k}(\mb y|\mb x)=p_{\theta_k}(\mb y|z_k=g_{\phi_k}(e_k=\text{Te}_{1:k}(\text{Emb}(\mb x))),\mb x), \label{eq:compeqsamp}
\end{align}
where $\text{Te}_{1:k}(\cdot)$ denotes the composite function $(\text{Te}_k\circ\text{Te}_{k-1}\circ\dotsb\circ\text{Te}_1)(\cdot)$.

$p_{\theta_k}(\mb y|z_k,\mb x)$ is modeled with a CDM based on an extension of the original denoising diffusion probabilistic model (DDPM) \cite{ho2020denoising} that includes additional covariates. For simplicity, the subscript $k$ in $\theta_k$ and $z_k$ is omitted, and a probability density function $p_{\theta}(\mb y|z,\mb x)$ is assumed. Here,  we consider a diffusion process that is fixed to a Markov chain with $T$ states, the joint probability given the covariates $z,\mb x$ and the response $\mb y$ is as follows \footnote{Here we denote $\mb y_{t=a}$ as $\mb y_a$ for simplicity, and $\mb y_0$ is equivalent to the response $\mb y$.}:
\begin{gather}
    q(\mb y_{1:T}\mid\mb y_0,z,\mb x)=\prod_{t=1}^Tq(\mb y_t\mid \mb y_{t-1},z,\mb x).
\end{gather}

The parametric form of the forward transition density function is represented as a Gaussian density function with a static variance schedule $\{\beta_1,\beta_2,\dotsc,\beta_T\}$. $\alpha_t:=1-\beta_t$ such that:
\begin{align}
    q(\mb y_t|\mb y_{t-1}, z, \mb x):=\;&\mathcal{N}(\mb y_t;\sqrt{\alpha_t}\mb y_{t-1}\notag\\&+\left(1-\sqrt{\alpha_t} \right)(z+\text{Enc}(\mb x)), \beta_t\mathbf{I}).
\end{align}
The encoder $\text{Enc}(\cdot)$ maps the image $\mb x$ to an embedding with the same dimensionality as $z$. Later sections will provide more details for the CDM, specifically pertaining to inference and sampling.

\begin{figure*}
  \centering
  \includegraphics[width=1\linewidth]{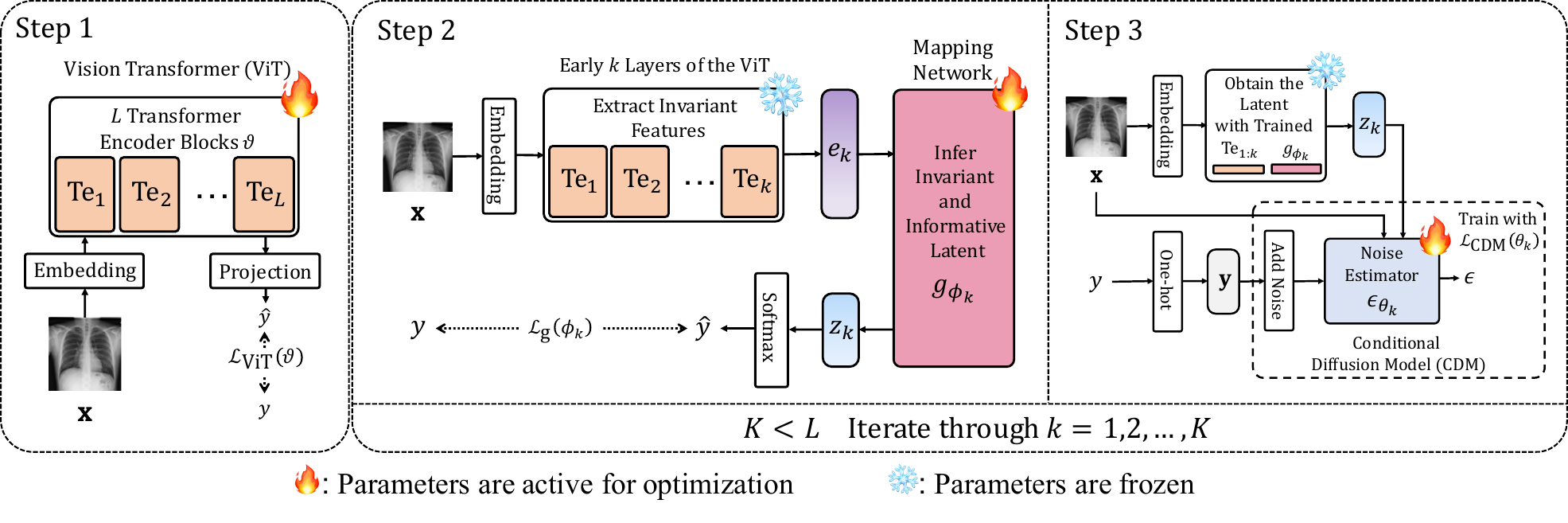}
  \caption{An illustration of the 3-step training procedure. Note that the training data point $\langle \mb x, y\rangle$ is sampled from $p_\text{train}$. In Step 1, the Vision Transformer (ViT) is trained to estimate its parameters $\vartheta$ in an end-to-end fashion using a cross-entropy loss $\mathcal{L}_\text{ViT}(\vartheta)$ (see \cref{eq:lossvit}). In Step 2, the ViT is frozen and the embedding $e_k$ is produced from the $k$-th transformer encoder block (in orange). The extracted embedding $e_k$ is passed through the mapping network (in red, with parameters $\phi_k$) to produce the latent $z_k$. The mapping network is trained by minimizing a cross-entropy loss $\mathcal{L}_\text{g}(\phi_k)$ (see \cref{eq:lossffn}) with the ground-truth label $y$ and the softmax-ed $z_k$. In Step 3, all transformer encoder blocks and the mapping network are frozen. The diffusion model is trained with parameters $\theta_k$ conditioned on $\mb x$ and $z_k$ to predict the noise term, and thus predict the denoised $\mb y$. This diffusion model is trained with the simplified objective $\mathcal{L}_\text{CDM}(\theta_k)$ (equation \cref{eq:losscdm}). The system iterates $k$ from $1$ to $K$ in Step 2 and Step 3, resulting in a total of $K$ ensemble members.}
  \label{fig:training}
\end{figure*}

\subsection{Training}
\label{subsec:training}
This section describes the process of estimating the parameters of the predictive density $p(\mb y|\mb x,\Theta)$ during training. Note that the mixture weight $\pi$ does not need to be estimated as it is set to be uniform. In order to estimate the parameters of the TE blocks, $\theta_k,\phi_k$, a 3-step training procedure is followed as illustrated in~\cref{fig:training}. Steps 2 and 3 are performed for $k=1,2,\dotsc,K$, so as to train $K$ ensemble members.

\textbf{Step 1 Training TE Blocks.}\; The parameters of the TE blocks are learned during the training of the ViT.
As discussed previously, the input $\mathbf{x}$ is initially transformed into an embedding and subsequently processed through $L$ Transformer Encoder (TE) blocks. The class token generated by the final TE block is used for classifying $\mathbf{x}$. For simpler notation, we denote a function $\text{Vit}:X\rightarrow Y$ to summarize all computations involved in the ViT including the last softmax function, where $\mb x\in X$ and $y\in Y$. The parameters of the ViT are estimated using maximum likelihood estimation (MLE), or equivalently, through minimizing the cross entropy (CEloss) between the prediction and the ground truth:
\begin{gather}
\mathcal{L}_{\text{ViT}}(\vartheta) := \Expect{\langle \mb x,y \rangle}{\text{CEloss}(y,\text{Vit}_\vartheta(\mb x))}. \label{eq:lossvit}
\end{gather}

\textbf{Step 2 Training Mapping Network.}\; Next, the parameters $\phi_k$ of the mapping network $g(\cdot)$ are estimated. In the proposed model, the latent variable $z$ serves as a conditioning signal in the diffusion process. $z$ is required to (i) provide relevant information about the ground truth label to facilitate the estimation of the predictive distribution (which is estimated by the CDM), while (ii) be robust to covariate shifts. Recall that the mediator $e\in\{e_{1:K}\}$ encodes the image into an embedding space that is insensitive to image distribution shift (under the constraint shown in~\cref{fig:l2}). Thus, we identify the latent $z$ as a non-linear transformation of the mediator $e$ (realised by a FFN $g(\cdot)$, see~\cref{eq:g}). Specifically, it represents the unnormalized probability (logit) when optimizing the cross-entropy loss with respect to the parameter $\phi_k$:
\begin{gather}
    \mathcal{L}_\text{g}(\phi_k):=\Expect{\langle e, y\rangle}{\text{CEloss}(y,\text{softmax}(z_k=g_{\phi_k}(e_k)))}. \label{eq:lossffn}
\end{gather}

\textbf{Step 3 Training Conditional Diffusion Model.}\; In the final step, the parameter $\theta_k$ of the noise estimator is learned in the conditional diffusion model. For simpler subscript notation, $k$ is omitted in $\theta_k$ and $z_k$ here. The parameter $\theta$ parameterizes the probability density function $p_\theta(\mb y|z,\mb x)$, which is estimated via minimizing the variational bound on the negative logarithmic likelihood (VBNL) of the conditional distribution $p_\theta(\mb y|z,\mb x)$:
\vspace{-1cm}
\begin{align}
    \mathcal{L}_\text{VBNL}(\theta)&:=\mathbb{E}[-\log p_\theta(\mb y|z,\mb x)]\leq\Expect{q}{\frac{p_\theta(\mb y_{0:T}|z,\mb x)}{q(\mb y_{1:T}| \mb y_0,z,\mb x)}} \notag \\ 
    &=\Expect{q}{\mathcal{L}_0+\sum_{t=2}^T\mathcal{L}_{t-1}(t)+\mathcal{L}_T}, \label{eq:vbnl}
\end{align}
where we have:
\begin{align}
    \mathcal{L}_0&:=-\underbrace{\log p_\theta(\mb y_0|\mb y_1,z,\mb x))}_\text{reconstruction term},\\
    \mathcal{L}_T&:=\underbrace{\infdiv[\big]{q(\mb y_T| \mb y_0,z,\mb x)}{p(\mb y_T|z,\mb x)}}_\text{prior matching term},\\
    \mathcal{L}_{t-1}(t)&:=\underbrace{\infdiv[\big]{q(\mb y_{t-1}|\mb y_t,\mb y_0,z,\mb x)}{p_\theta(\mb y_{t-1}|\mb y_t,z,\mb x)}}_\text{consistency term}.
\end{align}
Similar to the evidence lower bound in DDPM, this VBNL bound can be interpreted as three terms: (i) the reconstruction term, (ii) the prior matching term, and (iii) the consistency term. Among these, the prior matching term ($\mathcal{L}_T$) does not depend on any parameter and thus can be omitted during optimization.

In practice, a simplified variant of the VBNL is used for optimization (akin to the simplified objective in DDPM). Here, the noise term $\epsilon\sim\mathcal{N}(\mb 0,\mb I)$ is estimated to define $\mb y_t$ from $\mb y_0$ for $t\sim\text{Uniform}(1,2,\dotsc,T)$. Note that here we again omit the subscript $k$ in $\theta_k$ and $z_k$ for simplicity:
\begin{align}
    \mathcal{L}_\text{CDM}(\theta):=\Expect{\langle\mb x, \mb y_0\rangle,\epsilon,t}{\| \epsilon-\epsilon_\theta(\mb y_t,z,\mb x,t) \|_2^2}. \label{eq:losscdm}
\end{align}
$\mb y_t$ is computed by applying the reparameterization trick to the forward transition density function conditioned on $\mb y_0$, that is, $q(\mb y_t|\mb y_0,z,\mb x)$. Its functional form can be computed from the reparameterized $q(\mb y_t|\mb y_{t-1},z,\mb x)$ in a recursive fashion. The derived $\mb y_t$ is:
\begin{gather}
    \mb y_{t}=\:\sqrt{\Bar{\alpha}_t}\mb y_{0}+(1-\sqrt{\Bar{\alpha}_t})(z+\text{Enc}(\mb x))+\sqrt{1-\Bar{\alpha}_t}\epsilon,
\end{gather}
where $\bar{\alpha}_t:=\prod_{i=1}^t\alpha_i$.

\begin{algorithm*}
\caption{Drawing a sample from the CDM}
\label{alg:cdm_samp}
\begin{algorithmic}[1]
    \REQUIRE Image input \(\mb x\), latent variable \(z\), and learned parameter \(\theta\) including \( \text{Enc}(\cdot) \)
    \ENSURE A sample \( \mb y \) given \( \mb x \) and \( z \)
    \STATE Draw \( \mb y_{T}\sim \mathcal{N}(z,\mathbf{I}) \)
    \FOR{\( t \) in \( \{T,T-1,\dotsc ,1\} \)}
        \STATE Compute \( \Tilde{\mb y}_{0}=\frac{1}{\sqrt{\Bar{\alpha}_t}}\Big(\mb y_{t}-(1-\sqrt{\Bar{\alpha}_t})(z+\text{Enc}(\mb x))-\sqrt{1-\Bar{\alpha}_t}\epsilon_\theta(\mb y_t,z,\mb x,t)\Big) \)
        \IF{\(t>1\)}
            \STATE Draw \( \epsilon \sim \mathcal{N}(\mathbf{0}, \mathbf{I})\)
            \STATE Compute \( \mb y_{t-1}=\frac{\sqrt{\alpha_t}(1-\Bar{\alpha}_{t-1})}{1-\Bar{\alpha}_t}\mb y_{t}+\frac{\beta_t\sqrt{\Bar{\alpha}_{t-1}}}{1-\Bar{\alpha}_t}\Tilde{\mb y}_{0}-\left(\frac{\sqrt{\alpha_t}(1-\Bar{\alpha}_{t-1})+\beta_t\sqrt{\Bar{\alpha}_{t-1}}}{1-\Bar{\alpha}_t}-1\right)(z+\text{Enc}(\mb x))+\sqrt{\frac{\beta_t(1-\Bar{\alpha}_{t-1})}{1-\Bar{\alpha}_t}}\epsilon \)
        \ENDIF
    \ENDFOR
    \STATE Let \(\mb y=\mb y_0\)
    \RETURN \(\mb y\)
\end{algorithmic}
\end{algorithm*}

\subsection{Test-time Predictions}
\label{subsec:prediction}
In this subsection, the probabilistic model proposed in \cref{subsec:ppm} is used to predict the class given an image $\mb x$ sampled from $p_\text{test}(\mb x)$. Once trained, a 3-phase procedure is proposed to obtain the final prediction as illustrated in \cref{fig:prediction} (a).

Formally, predicting the response given the proposed mixture model $p(\mb y|\mb x, \Theta)$ is defined as follows: Given an image input $\mb x$, the class label is predicted by computing the conditional expectation $\mathbb{E}[\mb y|\mb x]$. Note that each component density $\mb p_k(\mb y|\mb x)$ does not have a trivial closed form. However, the reverse diffusion process allows us to sample from it. In practice, the expectation is computed by using the Monte Carlo (MC) method due to the intractability of $\int \mb y p(\mb y|\mb x, \Theta)\:d\mb y$:
\begin{align}
    \mathbb{E}[\mb y|\mb x]&=\int \mb y p(\mb y|\mb x, \Theta)\:d\mb y \\
    &=\int \mb y K^{-1} \sum_{k=1}^K\mb p_k(\mb y|\mb x) \:d\mb y \\
    &\approx (MK)^{-1}\sum_{m=1}^M\sum_{k=1}^K {\mb y}_{k,m}, \quad {\mb y}_{k,m}\sim\mb p_k(\mb y|\mb x),
\end{align}
where $M\in\intset^+$ should be as large as possible to achieve an accurate estimation.

To estimate the expected response (or class), one needs to sample ${\mb y}_{m,k}$ from the predictive component distribution $\mb p_k(\mb y|\mb x)$. As indicated in~\cref{eq:compeqsamp}, the component distribution is equivalent to the conditional distribution estimated by the CDM. To this end, a 3-phase procedure is proposed in order to sample from the CDM, and to estimate the expected response.

\textbf{Phase 1.}\; The value of the latent variable $z_k$ given the input $\mb x$ is computed through the trained transformer encoders and the mapping network. The computed value will be used as an informative and invariant conditioning signal for the CDM.

\textbf{Phase 2.}\; Once $z_k$ is computed, $M$ samples are drawn from the CDM's probability density function, $p_{\theta_k}(\mb y|z_k,\mb x)$. Each sample is seen as a candidate for the final prediction. In this work, drawing a sample from the CDM is performed by first drawing a noisy sample from the Gaussian prior and then gradually denoising it through the reverse diffusion process to obtain a clean sample. (For ease of reading, the subscripts for $\theta_k$ and $z_k$, and in $\mb y_{m,k}$ are omitted.) If $\theta$ is properly modeled, the consistency term $\mathcal{L}_{t-1}(t)$ is minimized. In the consistency term, the posterior of the forward transition density function is derived as:
\begin{align}
    q(\mb y_{t-1}|\mb y_t,\mb y_0,z,\mb x)=\mathcal{N}(\mb y_{t-1};\mu_q(\mb y_t,\mb y_0,z,\mb x),\Sigma_q(t)),
\end{align}
where its parameters are:
\begin{align}
    &\mu_q(\mb y_t,\mb y_0,z,\mb x)=\frac{\sqrt{\alpha_t}(1-\Bar{\alpha}_{t-1})}{1-\Bar{\alpha}_t}\mb y_{t}+\frac{\beta_t\sqrt{\Bar{\alpha}_{t-1}}}{1-\Bar{\alpha}_t}\mb y_{0} \notag \\
    &\;\;-\left(\frac{\sqrt{\alpha_t}(1-\Bar{\alpha}_{t-1})+\beta_t\sqrt{\Bar{\alpha}_{t-1}}}{1-\Bar{\alpha}_t}-1\right)(z+\text{Enc}(\mb x)),
\end{align}
and
\begin{align}
    \Sigma_q(t)=\sigma^2_q(t)\mb I=\frac{\beta_t(1-\Bar{\alpha}_{t-1})}{1-\Bar{\alpha}_t}\mathbf{I}.
\end{align}
Applying the reparameterization trick yields the following:
\begin{gather}
    \mb y_{t-1}=\mu_q(\mb y_t,\mb y_0,z,\mb x)+\sigma_q(t)\epsilon.
\end{gather}
To calculate $\mb y_{t-1}$, we require the value of $\mb y_0$ estimated at time step $t$. Recalling the forward transition density function at an arbitrary time step $q(\mb y_t|\mb y_0,z,\mb x)$, $\mb y_0$ given $\mb y_t$ is estimated as:
\begin{align}
    \mb y_{0}=\frac{1}{\sqrt{\Bar{\alpha}_t}}\Big(\mb y_{t}-(1-\sqrt{\Bar{\alpha}_t})(z+\text{Enc}(\mb x))-\sqrt{1-\Bar{\alpha}_t}\epsilon_\theta\Big).
\end{align}
A step-by-step algorithm summarizing the entire CDM sampling procedure can be found in \cref{alg:cdm_samp}.

Recall that the specification of the component $\mb p_k(\mb y|\mb x)$ enables us to draw a sample from its distribution via sampling from the CDM. Specifically, $\mb y\sim\mb p_k(\mb y|\mb x)$ are drawn through:
\begin{gather}
    e_k=\text{Te}_{1:k}(\text{Emb}(\mb x)),\; z_k=g_{\phi_k}(e_k),\; \mb y\sim p_{\theta_k}(\mb y|z=z_k,\mb x).
\end{gather}

\textbf{Phase 3.}\; Iterating $k$ from $1$ to $K$ results in a set of $K\times M$ samples. Note that since each sample is a $A$-dimensional vector ($A$ is the number of classes) rather than a scalar class label. Therefore, each sample is mapped to a probability simplex after averaging them.

\textbf{Mapping Sample Space to Probability Simplex.}\; The CDM guides the mixture model to treat $\mb y$ as a vector sampled from a real-valued set rather than a categorical distribution. This is due to the fact that, in the context of denoising score matching, the loss function used during CDM's optimization effectively becomes the squared error (i.e. the Brier score) between the estimated denoised $\mb y$ and the actual clean $\mb y^*$ from the data distribution~\cite{luo2022understanding,karras2022elucidating}. Consequently, it is crucial to map the sampled $\mb y$ onto the probability simplex.

Given the estimated expected response vector $\mb y \in \mathbb{R}^A$  obtained by averaging all $K \times M$ samples from the mixture model $p(\mb y|\mb x, \Theta)$, let $\mb y^a$ represent its value in the $a$-th dimension. The probability of the final prediction being the class $a$ is then calculated in the weighted-softmax form of the Brier score \cite{brier1950verification}. Specifically, we follow the mapping function introduced in diffusion-based classifiers by Han \textit{et al.} \cite{han2022card}, where the hyperparameter $\iota$ controls the sharpness of the probability distribution:
\begin{align}
    \Pr(y=a|\mb x)=\frac{\mathrm{exp}\left(-\iota^{-1}(\mb y^a-1)^2\right)}{\sum_{i=1}^A \mathrm{exp}\left(-\iota^{-1}(\mb y^i-1)^2\right)}.
\end{align}

\section{Experiments and Results}
\label{sec:exp}

\begin{figure}
  \centering
  \includegraphics[width=0.9\linewidth]{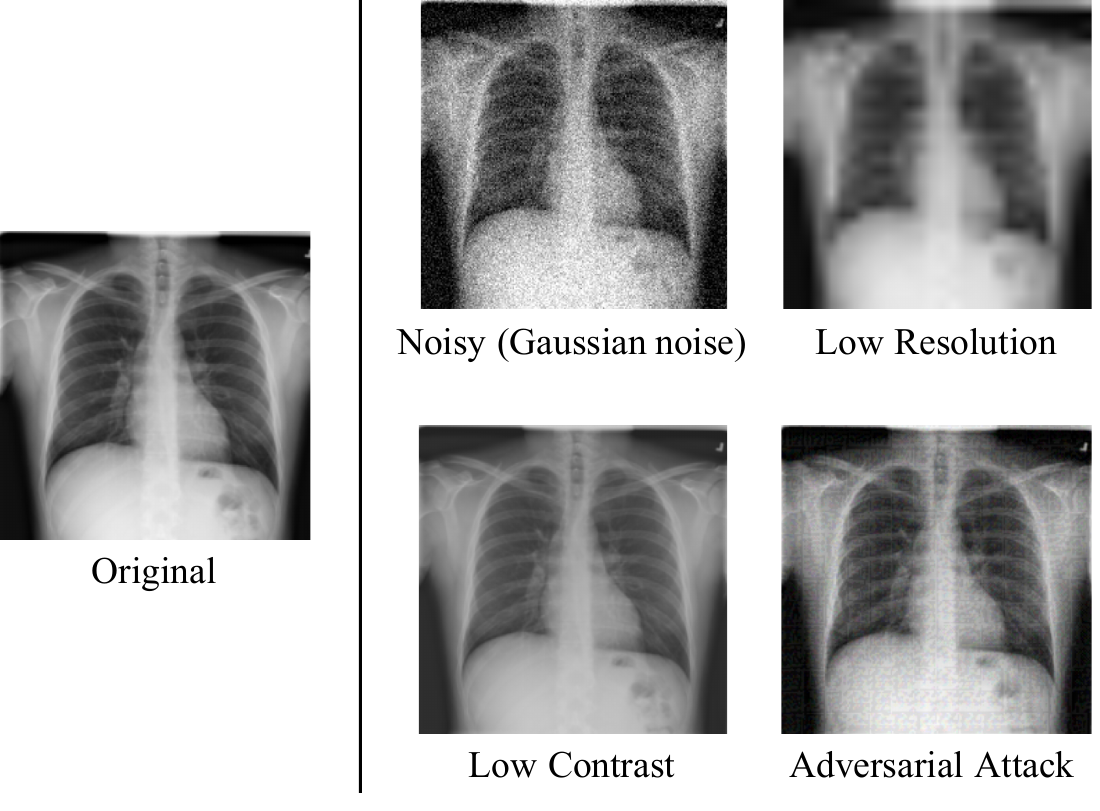}
  \caption{Illustration of the Tuberculosis chest X-ray dataset under different perturbations (noisy $\mathbb{g}=0.10$, low resolution $w=8.00$, contrast $r=0.70$, FGSM adversarial attack $\varepsilon=0.03$).}
  \label{fig:data}
\end{figure}

We evaluate the proposed method on two medical imaging benchmarking datasets: Tuberculosis chest X-ray dataset~\cite{rahman2020reliable} and the ISIC Melanoma skin cancer dataset~\cite{rotemberg2021patient}. The Tuberculosis chest X-ray dataset consists of X-ray images of 3500 patients with Tuberculosis and 3500 patients without Tuberculosis. The ISIC Melanoma skin cancer dataset contains lesion images of 5085 patients with malignant skin cancer and 5480 patients with benign tumor. A range of baseline methods are chosen for comparisons and these cover a variety of different architectures:
\begin{itemize}[labelwidth=!, topsep=0pt, leftmargin=*]
    \item \textbf{Comparison with Non-ensemble Methods:}\; To evaluate the advantages of the ensembling framework over using individual models, several widely used non-ensemble methods are included: ResNets \cite{he2016deep} and ViTs \cite{dosovitskiyimage}. Additionally, comparisons with models based on hybrid architectures are included, such as MedViT~\cite{manzari2023medvit}.
    \item \textbf{Comparison with Existing Ensemble Methods:}\; To provide a comprehensive evaluation, the proposed ensemble method is compared against state-of-the-art ensemble deep learning methods specifically designed for medical image classification, including deep tree training of convolutional ensembles (DTT)~\cite{yang2021two},  improved convolutional ensembles (ICNN-Ensemble)~\cite{musaev2023icnn} and dynamic-weighted ensembles (DWE)~\cite{Pacheco2020Learning}.
\end{itemize}

\textbf{Experiment Configuration Details.}\; For the chest X-ray dataset, the split for the image-label pairs in training/validation/testing set is $5670/630/700$. For the ISIC skin cancer dataset, the split for the image-label pairs in training/validation/testing set is $7605/1000/1960$. In both datasets, all images have binary labels. The test sets are balanced. Several empirical choices were made: 5 mixture components ($K=5$), and each were sampled 20 times ($M=20$). For the diffusion model, 1000 time-steps ($T=1000$) were chosen, with a noise schedule of $\beta_1=10^{-4}$, $\beta_T=0.02$. For the probability simplex mapping, we determine the optimal value of $\iota$ by minimizing the ECE on the validation set, using the Nelder-Mead method \cite{nelder1965simplex}, specifically, $\iota = 0.1737$ was set for the X-ray dataset, and $\iota= 0.3162$ for the skin cancer dataset. The transformer encoder blocks in LaDiNE are extracted from ViT-B~\cite{dosovitskiyimage}, and all mapping networks are implemented by multilayer perceptrons (MLPs) with 3 hidden layers. All CDMs and $\text{Enc}(\cdot)$ are implemented with MLPs with 3 hidden layers, the conditioning in the CDM (time and image $\mathbf{x}$ conditioning) is implemented with element-wise product. All baseline models and LaDiNE were trained from scratch until loss convergence. For the baseline methods that require selecting ensemble members, the procedures in the original papers were followed exactly as described. The implementation of LaDiNE, including the code for models as well as training/evaluation scripts, is made publicly available\footnote{The code repository: https://github.com/xingbpshen/nested-diffusion}.

\subsection{Results without Perturbations} When no perturbations are performed on the input images, performance is very good for all methods. For the chest X-ray dataset, when presented with clean inputs (i.e., without simulated covariate shifts), all methods achieve classification accuracies that exceed $99.00\%$. For the skin cancer dataset, the proposed method achieves accuracies of $94.18\%$ on clean inputs, on par with the best-performing method ResNet-18 which attains accuracies of $95.02\%$\footnote{For the clean chest X-ray dataset, ViT-B achieves $99.75\%$ accuracy, LaDiNE achieves $99.90\%$ accuracy. For the clean skin cancer dataset, ViT-B achieves $92.93\%$ accuracy.}.

\subsection{Results Under Perturbations}
The robustness of LaDiNE against competing methods is examined by providing the network with images at test time that have been perturbed in ways that were not previously seen during training. To simulate significant covariate shifts, a variety of perturbations were performed on the clean test set images. These perturbations included adding Gaussian noise, altering image resolution, and adjusting contrast levels. Specifically, the following transformation functions are defined:
\begin{itemize}[labelwidth=!, topsep=0pt, leftmargin=*]
    \item \textbf{Gaussian Noise.}\; The function $\mathcal{T}_\text{gn}:\realset^{H\times W\times C}\rightarrow\realset^{H\times W\times C}$ is defined such that 
    \begin{align}
        \mathcal{T}_\text{gn}(\mb x;\mathbb{g}):=\mb x+\mathbb{g}\epsilon, 
    \end{align}
    where $\mb x$ is an image in the test set, and $\epsilon\sim\mathcal{N}(\mathbf{0},\mathbf{I})$. $\mathbb{g}\in\realset$ is a scalar that controls the scale of the injected noise.
    \item \textbf{Low Image Resolution.}\; Reduced-resolution images are produced by defining a function $\mathcal{T}_\text{lr}:\realset^{H\times W\times C}\rightarrow\realset^{H\times W\times C}$ that
    \begin{align}
        \mathcal{T}_\text{lr}(\mb x;w):=\text{Resize}(\text{DownSample}(\mb x, w), (H, W, C)),
    \end{align}
    where $\mb x$ is an image in the test set, and $w$ is the downsampling factor. This function reduces the resolution of the image and then resizes it back to the original dimensions, simulating a low-resolution effect.
    \item \textbf{Image Color Contrast.}\; The color contrast of the images are manipulated through the function $\mathcal{T}_\text{c}:\realset^{H\times W\times C}\rightarrow\realset^{H\times W\times C}$ that
    \begin{align}
        \mathcal{T}_\text{c}(\mb x;r):=r(\mb x-\bar{\mb x})+\bar{\mb x},
    \end{align}
    where $\mb x$ is an image in the test set, $\bar{\mb x}$ is the mean value of all pixels in the image $\mb x$ for each channel, and $r\in\realset$ is a scalar that controls the contrast level. This function adjusts the contrast of the image by scaling the variance of the pixel values, enhancing or reducing the overall contrast according to the value of $r$.
\end{itemize}

\begin{table*}
\centering
\caption{Comparison in classification accuracy (\%) with state-of-the-art methods on two benchmark datasets with unseen input perturbations. Methods are categorized into four groups, from top to bottom: (i) Models based on convolutional neural networks (CNNs). (ii) Models based on Transformers. (iiI) hybrid models with CNNs and Transformers (ConViT-B, MedViT-B), and diffusion models (CARD). (iv) Ensemble learning methods.}
\tabcolsep=0.092cm
\label{tab:acccomb}
\begin{tabular}{l|cccc|cccc}
\toprule
\multirow{3}{*}{Methods} & \multicolumn{4}{c|}{Chest X-ray}                                & \multicolumn{4}{c}{ISIC}                                       \\ \cmidrule{2-9} 
                         & \multicolumn{2}{c}{Gaussian noise} & Low resolution & Contrast & \multicolumn{2}{c}{Gaussian noise} & Low resolution & Contrast \\
                         & $\mathbb{g}=0.50$             & $\mathbb{g}=1.00$            & $w=4.00$           & $r=0.70$     & $\mathbb{g}=0.50$             & $\mathbb{g}=1.00$            & $w=4.00$           & $r=0.70$     \\ \midrule
ResNet-18 \cite{he2016deep}               & $50.00\pm0.00$ & $50.00\pm0.00$  & $50.00\pm0.00$  & $99.57\pm0.11$ & $50.56\pm2.23$ & $49.64\pm0.94$ & $81.48\pm2.88$ & $92.37\pm0.28$ \\
ResNet-50 \cite{he2016deep}               &  $50.00\pm0.00$ & $50.38\pm0.54$  & $50.00\pm0.00$ & $\mb{99.86}\pm\mb{0.20}$  & $51.02\pm0.00$ & $54.34\pm4.69$ & $80.48\pm3.40$ & $92.42\pm0.09$         \\
EfficientNetV2-L \cite{tan2021efficientnetv2}        & $50.00\pm0.00$ & $50.00\pm0.00$ & $96.86\pm0.42$ & $93.10\pm0.70$ &  $48.98\pm0.00$  &  $48.98\pm0.00$  &  $89.73\pm0.31$  & $88.45\pm1.50$    \\ \midrule
DeiT-B \cite{touvron2021training}                  & $68.86\pm5.36$ &  $57.57\pm6.25$ & $94.43\pm2.89$ & $99.57\pm0.00$ &  $69.32\pm2.04$ &  $64.69\pm1.92$  & $87.75\pm0.15$  &  $92.54\pm0.67$      \\
ViT-B \cite{dosovitskiyimage}                   & $74.34\pm2.89$ & $57.76\pm4.18$ &  $94.71\pm0.71$ &  $97.14\pm0.12$ & $71.90\pm8.86$  & $55.42\pm4.00$  &  $89.72\pm0.37$ & $91.58\pm0.40$         \\
Swin-B \cite{liu2021swin}                  &  $59.81\pm0.04$ &  $50.00\pm0.00$ & $59.29\pm3.25$  & $98.29\pm2.42$ &  $67.43\pm1.72$ &  $63.93\pm0.83$ & $88.44\pm0.60$ & $91.34\pm0.25$         \\ \midrule
ConViT-B \cite{d2021convit}                &  $76.57\pm4.69$ & $55.00\pm2.53$ &  $94.62\pm1.67$  &  $99.33\pm0.29$ &  $70.86\pm1.69$ &  $60.83\pm4.08$  &  $90.62\pm0.90$   &  $92.93\pm0.04$        \\
MedViT-B \cite{manzari2023medvit}                &  $73.00\pm3.05$  &  $52.95\pm2.20$  & $96.24\pm0.57$  &  $94.67\pm0.49$ &  $61.63\pm2.02$  &  $47.58\pm2.15$  &  $91.46\pm0.46$  &  $90.92\pm0.37$       \\
CARD \cite{han2022card} & $75.38\pm2.86$ & $57.79\pm3.25$ & $94.86\pm0.65$ & $97.95\pm0.13$ & $72.06\pm8.25$ & $55.41\pm3.67$ & $90.20\pm0.34$ & $91.80\pm0.36$ \\ \midrule
Deep Ensembles \cite{lakshminarayanan2017simple} & $50.00\pm0.00$ & $50.00\pm0.00$ & $83.86\pm0.23$ & $92.71\pm0.31$ & $50.00\pm0.00$ & $50.00\pm0.00$ & $88.32\pm0.08$ & $91.09\pm0.06$ \\
DWE \cite{Pacheco2020Learning} & $74.14\pm0.58$ & $40.52\pm1.58$ & $68.14\pm0.12$ & $72.62\pm0.13$ & $58.18\pm0.23$ & $55.14\pm1.29$ & $71.02\pm0.07$ & $71.68\pm0.11$ \\
DTT \cite{yang2021two} & $75.67\pm0.64$ & $62.90\pm0.47$ & $97.71\pm0.42$ & $95.14\pm0.51$ & $50.60\pm0.09$ & $50.02\pm0.10$ & $\mb{92.52}\pm\mb{0.06}$ & $91.85\pm0.05$ \\
ICNN-Ensemble \cite{musaev2023icnn} & $75.86\pm0.42$ & $61.43\pm0.93$ & $97.05\pm0.41$ & $96.00\pm0.65$ & $50.34\pm0.15$ & $50.12\pm0.10$ & $87.57\pm0.30$ & $89.52\pm0.11$ \\
SEViT \cite{almalik2022self}                   & $69.19\pm3.27$  &  $62.24\pm4.36$  & $97.76\pm0.33$  & $97.15\pm0.20$ & $67.04\pm2.59$ &  $54.42\pm3.64$  &  $90.16\pm0.72$  &  $92.00\pm0.14$        \\
LaDiNE (proposed)               & $\mb{78.33}\pm\mb{0.69}$ &  $\mb{66.33}\pm\mb{2.07}$  &  $\mb{98.90}\pm\mb{0.49}$ & $97.86\pm0.23$ & $\mb{73.16}\pm\mb{2.66}$  & $\mb{69.93}\pm\mb{2.35}$  &  $91.17\pm0.36$  &  $\mb{93.14}\pm\mb{0.24}$        \\ \bottomrule
\end{tabular}
\end{table*}

\textbf{Results.}\; The experimental results for the robustness experiments on both datasets can be found in~\cref{tab:acccomb}. The results presented indicate the means and standard deviations of the classification accuracies (in percentages) over three runs. The overall trend indicates that LaDiNE consistently outperforms other methods across almost all perturbations, highlighting its effectiveness and robustness in handling noisy and perturbed images. In particular, LaDiNE shows superior performance on both datasets under high noise levels ($\mathbb{g}=1.00$), and demonstrates the highest robustness among all tested models overall. Traditional models such as ResNet-18 and ResNet-50 exhibit poor performance under Gaussian noise, with accuracies dropping to $50\%$ \footnote{Note that ResNet-50 consistently provides predictions of ``Healthy'' for all input images perturbed with Gaussian noise ($\mathbb{g}=1.00$).}, indicating a failure to generalize under noisy conditions.

When handling lower-resolution images, LaDiNE achieves the highest accuracies, particularly in the chest X-ray dataset ($98.90\%$). Although model performance typically declines with lower resolution input images, the transformer-based models (i.e. the second group of models in~\cref{tab:acccomb}), such as ViT-B and DeiT-B, maintain relatively high accuracies as compared to CNN-based models. When handling lower-contrast input images, LaDiNE and ResNet-50 lead in robustness on the ISIC dataset, with LaDiNE achieving $93.14\%$ accuracy. Transformer-based models generally show robust performance across varying contrast levels, with SEViT and ConViT-B also performing well.

\subsection{Results Under Adversarial Attacks}
Adversarial attacks can seriously compromise the reliability and safety of medical imaging models deployed in clinical settings. Several medical machine learning papers have illustrated how these attacks could result in incorrect diagnoses, inappropriate treatments, and even financial exploitation through insurance fraud~\cite{finlayson2019adversarial,kaviani2022adversarial,bortsova2021adversarial}.

Adversarial attacks can be formulated as a covariate shift context where the adversarially perturbed inputs $\mb{x}^\text{adv}$ are sampled from a distribution $p_\text{adv}(\mb{x}^\text{adv})$ that differs from the original distribution $p(\mb{x})$, while the conditional distribution of labels given the input remains unchanged, i.e., $p(y|\mb{x}) = p_\text{adv}(y|\mb{x}^\text{adv})$. In this setting, the adversarial attack induces a shift in the marginal distribution of the input, creating a context where the model encounters input distributions during testing (or deployment) that deviate from those it was trained on, yet where the underlying relationship between the input and the label remains consistent.

To complement existing studies on adversarial robustness in medical image classification, this work presents experimental results testing adversarial robustness of ensemble learning methods. Following the procedure described in~\cite{almalik2022self},  adversarial perturbation $\mathcal{T}_{\text{adv}}(\cdot)$ is applied to image $\mb x$ based on the gradient of the backbone model (e.g. ViT or ResNet) $\mathcal{M}_\text{base}(\cdot)$, such that $\| \mb x-\mathcal{T}_{\text{adv}}(\mb x) \|_\infty \leq \varepsilon$, where $\varepsilon\in\realset$ is a divergence threshold, and $\mathcal{M}_\text{base}(\mb x)\neq\mathcal{M}_\text{base}(\mathcal{T}_{\text{adv}}(\mb x))$. In this work, the top performing methods (for robustness against noisy conditions) are chosen in order to assess their robustness to adversarial attacks. For a maximally comprehensive assessment, three gradient-based methods are deployed to generate adversarial images with a threshold value of $\varepsilon=0.03$: (i) Fast Gradient Sign Method (FGSM)~\cite{goodfellow2014explaining}; (ii) Projected Gradient Descent (PGD)~\cite{madry2018towards}; (iii) Auto-PGD~\cite{croce2020reliable}.

\textbf{Results.}\; The results presented in~\cref{tab:accadv} demonstrate the performance of various methods under adversarial attacks using FGSM, PGD, and AutoPGD algorithms. LaDiNE, consistently outperforms other models across both datasets and all attack types, indicating superior robustness to adversarial attacks.

For the chest X-ray dataset, LaDiNE achieves the highest classification accuracy for FGSM, PGD, and AutoPGD attacks, respectively. In comparison, SEViT shows strong performance but falls short of LaDiNE, especially under PGD and AutoPGD attacks. This shortfall can be attributed to SEViT's reliance on the final prediction of the ViT, which is particularly vulnerable to adversarial perturbations (in contrast, LaDiNE does not rely on ViT's final prediction). Traditional deep learning models such as ResNet-50 and EfficientNetV2-L, as well as ensemble methods, perform poorly under these adversarial conditions, with accuracies often dropping to near zero under PGD and AutoPGD attacks.

For the skin cancer dataset, LaDiNE also leads with accuracies of $60.15\%$, $61.60\%$, and $61.30\%$ for FGSM, PGD, and AutoPGD attacks, respectively. While SEViT performs relatively well, with accuracy scores in the mid-50s, other methods like ViT-B and MedViT-B show vulnerabilities to adversarial perturbations, with accuracies dropping substantially under more sophisticated attacks like PGD and AutoPGD.

\begin{table*}
\centering
\caption{Comparison in classification accuracy (\%) with state-of-the-art methods on two benchmarks with adversarial attacks.}
\label{tab:accadv}
\begin{tabular}{l|ccc|ccc}
\toprule
\multirow{2}{*}{Methods} & \multicolumn{3}{c|}{Chest X-ray}          & \multicolumn{3}{c}{ISIC}                 \\ \cmidrule{2-7} 
                         & FGSM \cite{goodfellow2014explaining} & PGD \cite{madry2018towards} & AutoPGD \cite{croce2020reliable} & FGSM & PGD & AutoPGD \\ \midrule
ResNet-50 \cite{he2016deep}               & $46.72\pm3.40$     & $0.00\pm0.00$    & $0.00\pm0.00$                            & $55.32\pm0.24$     & $0.00\pm0.00$    & $0.00\pm0.00$                            \\
EfficientNetV2-L \cite{tan2021efficientnetv2}        & $34.28\pm1.02$     & $0.19\pm0.07$    & $0.14\pm0.12$                            & $19.75\pm6.24$     & $0.00\pm0.00$    & $0.00\pm0.00$                            \\
DeiT-B \cite{touvron2021training}                  & $35.28\pm1.76$     & $0.00\pm0.00$    & $0.00\pm0.00$                            & $26.33\pm7.15$     & $0.00\pm0.00$    & $0.00\pm0.00$                            \\
ViT-B \cite{dosovitskiyimage}                   & $15.38\pm3.68$     & $0.14\pm0.12$    & $0.00\pm0.00$                            & $22.20\pm8.22$     & $0.29\pm0.17$    & $0.00\pm0.00$                            \\
ConViT-B \cite{d2021convit}                & $20.52\pm2.66$     & $0.00\pm0.00$    & $0.00\pm0.00$                            & $35.95\pm0.74$     & $0.02\pm0.02$    & $0.00\pm0.00$                            \\
MedViT-B \cite{manzari2023medvit}                & $10.29\pm4.50$     & $2.95\pm2.10$    & $0.38\pm0.36$                            & $23.93\pm6.28$     & $0.00\pm0.00$    & $0.00\pm0.00$                            \\
CARD \cite{han2022card} & $14.81\pm2.01$     & $0.00\pm0.00$    & $0.00\pm0.00$                            & $18.50\pm4.13$     & $0.20\pm0.13$    & $0.00\pm0.00$ \\
Deep Ensembles \cite{lakshminarayanan2017simple} & $34.67\pm0.07$ & $0.00\pm0.00$ & $0.00\pm0.00$ & $20.71\pm0.07$ & $0.00\pm0.00$ & $0.00\pm0.00$ \\
DTT \cite{yang2021two} & $10.29\pm0.47$ & $0.76\pm0.07$ & $0.57\pm0.03$ & $26.46\pm0.06$ & $5.83\pm0.03$ & $2.48\pm0.02$ \\
ICNN-Ensemble \cite{musaev2023icnn} & $49.86\pm0.31$ & $0.00\pm0.00$ & $0.00\pm0.00$ & $41.46\pm0.28$ & $1.11\pm0.02$ & $0.00\pm0.00$ \\
SEViT \cite{almalik2022self}                   & $85.90\pm3.39$     & $92.76\pm0.86$    & $94.24\pm1.36$                            & $54.15\pm3.56$     & $51.52\pm4.99$    & $57.30\pm8.74$                            \\
LaDiNE (proposed)               & $\mb{94.86}\pm\mb{0.20}$     & $\mb{96.10}\pm\mb{0.94}$    & $\mb{96.05}\pm\mb{1.48}$                            & $\mb{60.15}\pm\mb{4.93}$     & $\mb{61.60}\pm\mb{4.53}$    & $\mb{61.30}\pm\mb{2.70}$                            \\ \bottomrule
\end{tabular}
\end{table*}

\subsection{Results on the Quality of Prediction Confidence}
In high-stakes domains such as clinical decision-making, it is crucial to assess whether a model’s predicted confidence aligns with its actual performance. One common metric for this is Expected Calibration Error (ECE), which measures the discrepancy between confidence scores and observed accuracy~\cite{naeini2015obtaining,guo2017calibration,jungo2020analyzing,shui2023mitigating}. In this work, we evaluate confidence calibration using ECE across various covariate shift scenarios. Proper calibration is vital in clinical settings to ensure the model's confidence reliably reflects its true performance, reducing the risk of over-confident and potentially erroneous predictions.

The ECE measures the weighted average of the differences between predicted confidence and accuracy, over all confidence levels. To compute ECE, the predictions are divided into several bins based on their confidence scores. For each bin, the accuracy and confidence are calculated, and the absolute difference between them is weighted by the number of samples in the bin. The formula for ECE with $b$ bins is given by:
\begin{align}
    \mathrm{ECE}_b:=\sum_{i=1}^{b}\frac{|B_i|}{u}\Big|\mathbf{acc}(B_i)-\mathbf{conf}(B_i)\Big|, \label{eq:ece}
\end{align}
where $B_i$ denotes the set of indices of predictions that fall into bin $i$, $u$ is the total number of predictions, $\mathbf{acc}(B_i)$ is the empirical accuracy for bin $i$, i.e., the fraction of correct predictions in the bin, and $\mathbf{conf}(B_i)$ is the average confidence score for bin $i$.

\begin{figure*}[]
  \centering
  \subfigure{\includegraphics[width=0.24\linewidth]{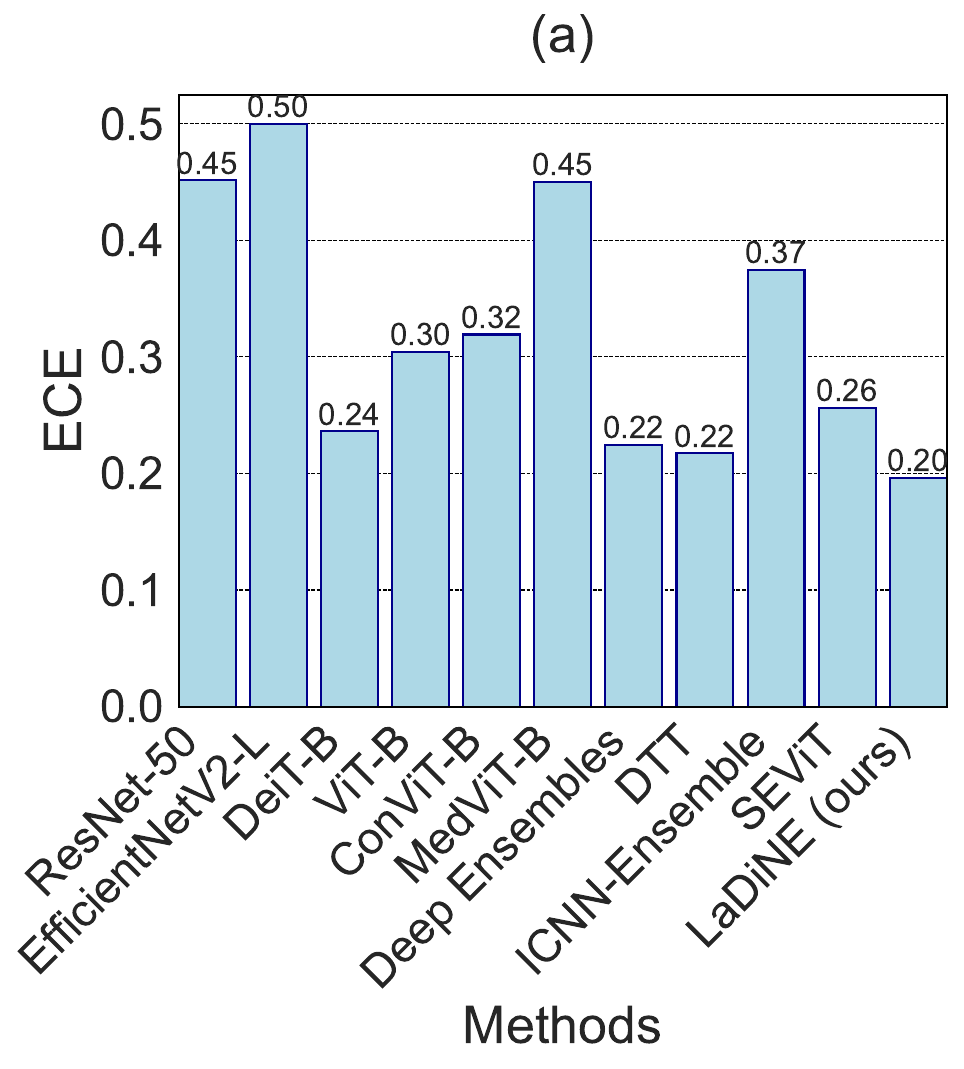}}
  \subfigure{\includegraphics[width=0.24\linewidth]{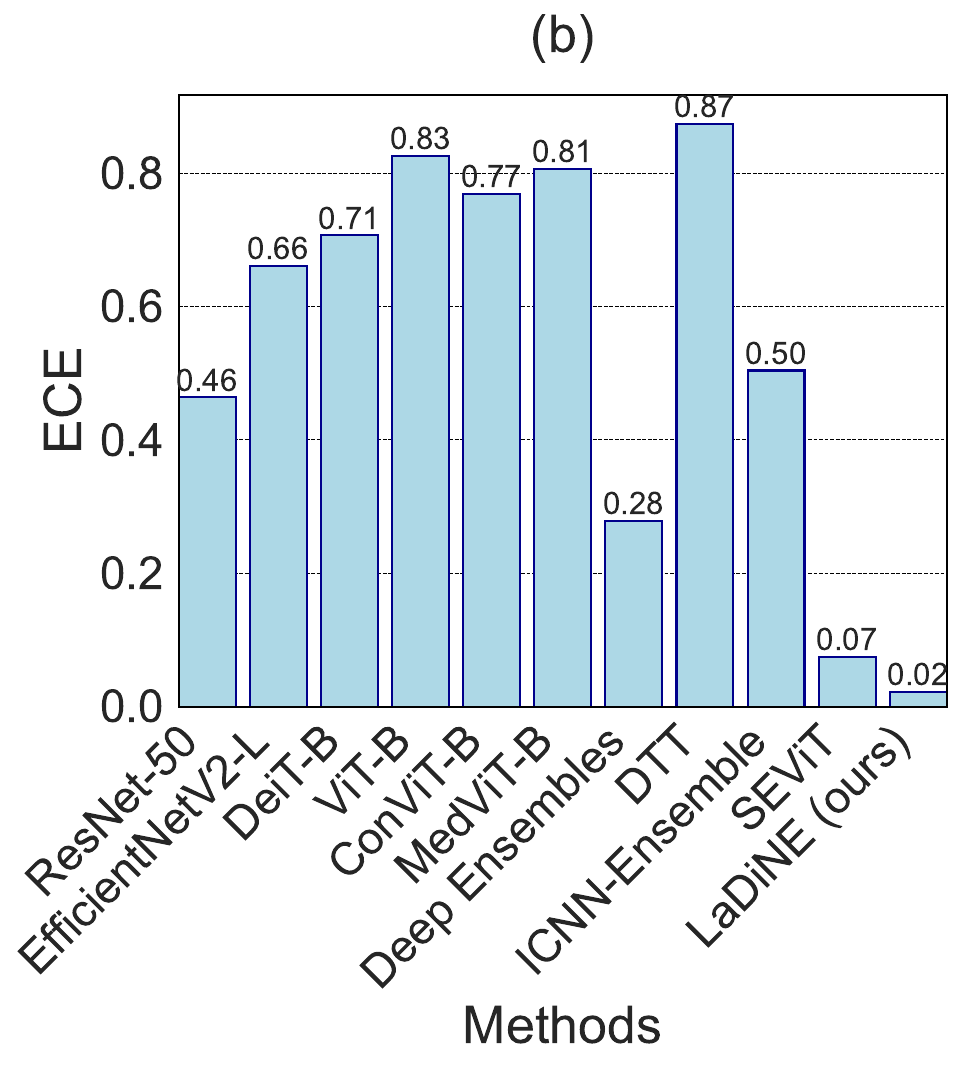}}
  \subfigure{\includegraphics[width=0.24\linewidth]{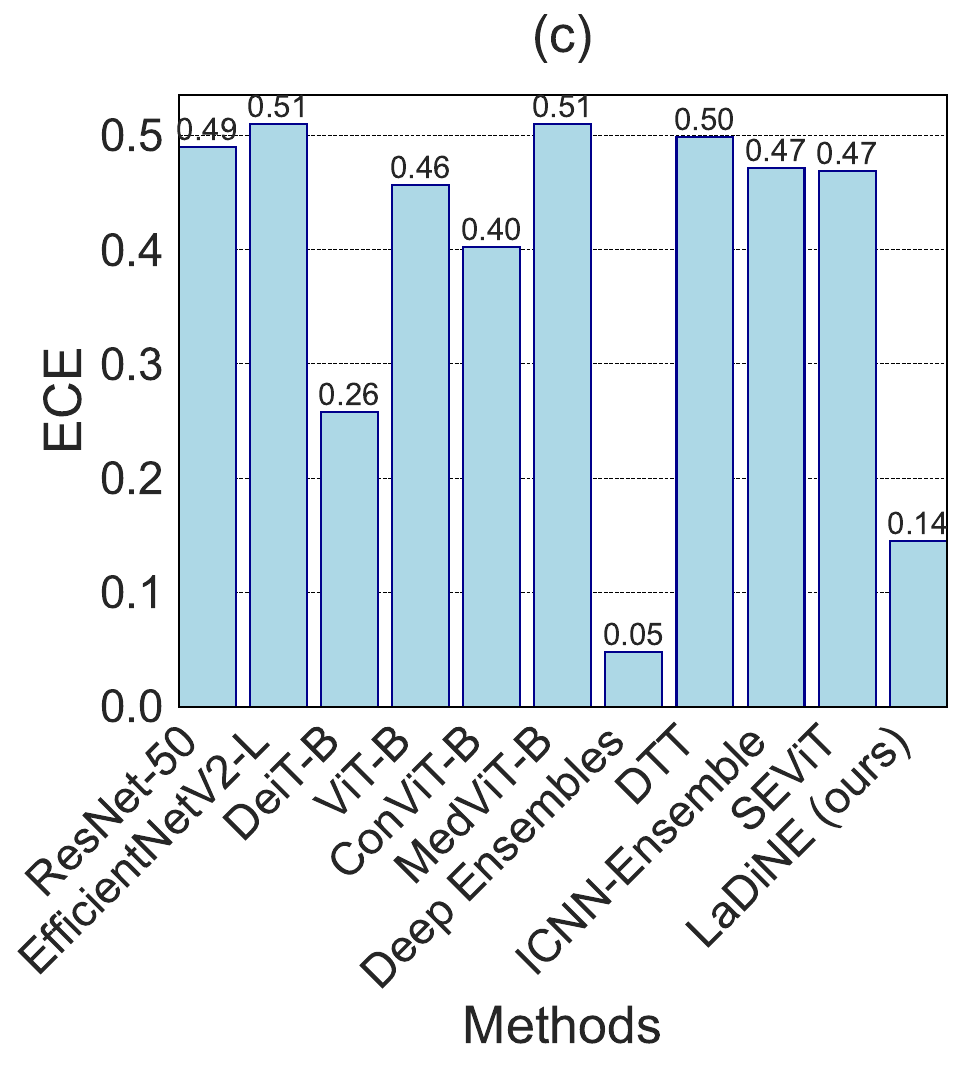}}
  \subfigure{\includegraphics[width=0.24\linewidth]{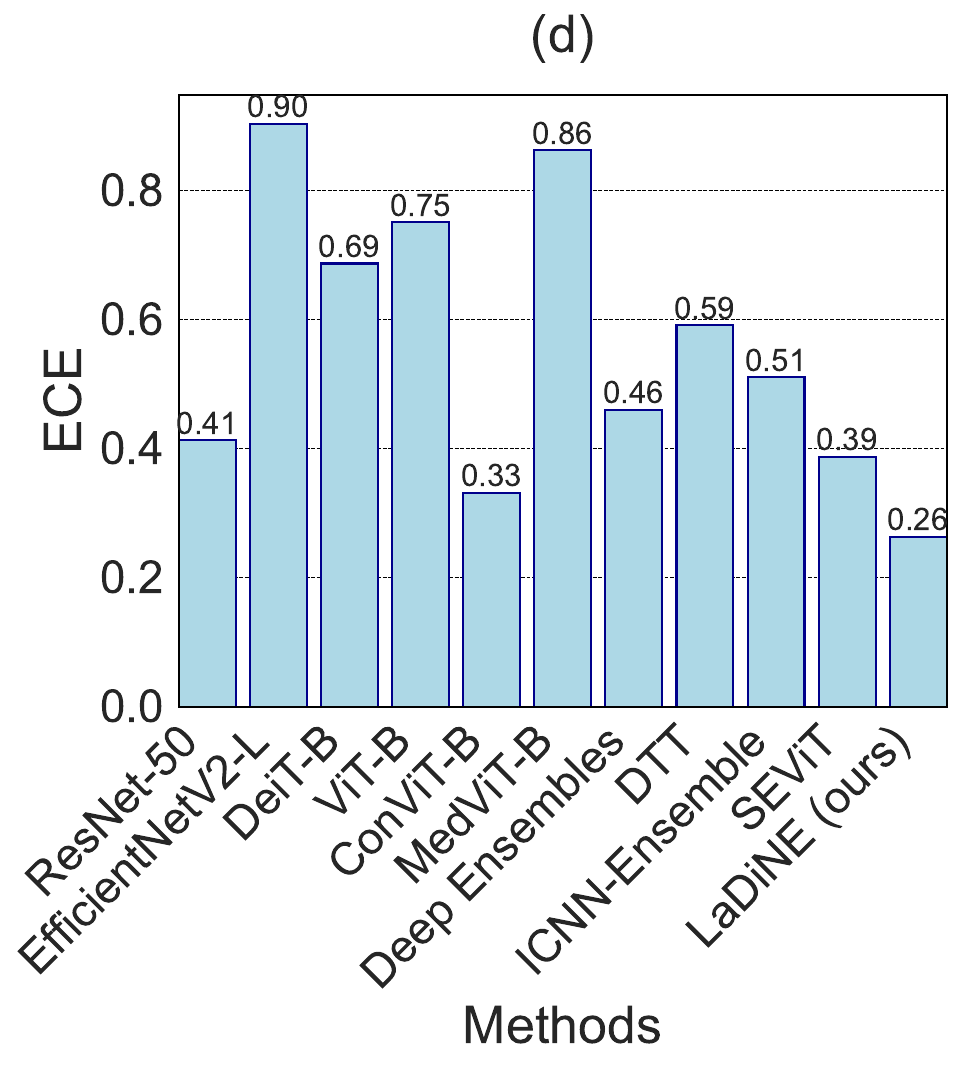}}
  \caption{Plot of expected calibration error (ECE) with a uniform set of ten bins: (a) Gaussian noise injection ($\mathbb{g}=1.00$) in the chest X-ray dataset. (b) FGSM attack ($\varepsilon=0.03$) in the chest X-ray dataset. (c) Gaussian noise injection ($\mathbb{g}=1.00$) in the skin cancer dataset, (d) FGSM attack ($\varepsilon=0.03$) for the skin cancer dataset.}
  \label{fig:ece}
\end{figure*}

\textbf{Results.}\; All methods are tested under covariate shifts on both datasets. Accurately expressing their prediction confidence is important in order to avoid being over-confident in incorrect predictions. Overall, LaDiNE and Deep Ensembles, a scalable method for improving predictive uncertainty estimation~\cite{lakshminarayanan2017simple}, show stronger capabilities in providing well-calibrated confidence scores among all methods in both datasets. Furthermore, LaDiNE achieves a lower ECE than Deep Ensembles (i) under adversarial perturbations in both datasets and (ii) under Gaussian noise injections in the chest X-ray dataset, as illustrated in~\cref{fig:ece}. Under Gaussian noise injections, Deep Ensembles reaches the lowest ECE in the skin cancer dataset, however, its classification accuracy under this condition is $50.00\%$ which indicates low predictive power as compared to LaDiNE's accuracy of $73.16\%$ in this case.

\subsection{Quantifying Instance-level Uncertainty}
Quantifying the reliability of a model’s predictions is critical in clinical settings, where the consequences of presenting incorrect predictions can be significant. In order to maintain trust in the system, the model should quantify the level of uncertainty in each prediction, with the goal of being correct when confident, and uncertain when incorrect~\cite{mehta2022qu}. In this fashion, clinicians can focus their review on cases where the model is less certain, thereby improving decision-making and fostering trust in the system. This section shows results in the quantification of LaDiNE's prediction uncertainties given an image instance under covariate shifts.

Recall that LaDiNE provides $K\times M$ prediction vectors $\hat{\mb y}$ for a given image $\mb x$ (see \cref{subsec:prediction}). These prediction vectors are denoted as a set $\mathcal{S}$ and a set of scalars are defined: $\mathcal{S}_a:=\{\hat{\mb y}_i^a\mid \hat{\mb y}_i\in\mathcal{S}\}$ (note that $\hat{\mb y}_i^a$ denotes the value of $\hat{\mb y}_i$ in the $a$-th dimension). Uncertainties are measured with respect to the consistency among the samples predictions. To this end, two methods are used:

\begin{itemize}[labelwidth=!, topsep=0pt, leftmargin=*]
    \item \textbf{Class-wise Prediction Interval Width (CPIW).}\; The CPIW, defined by Han \textit{et al.}, measures the uncertainties of the predictions provided by a diffusion model \cite{han2022card}:
    \begin{align}
        \mathrm{CPIW}_a:=\mathcal{Q}_{97.5}(\mathcal{S}_a)-\mathcal{Q}_{2.5}(\mathcal{S}_a),
    \end{align}
    where $\mathcal{Q}_n(\cdot)$ calculates the $n$-th percentile. CPIW measures the spread of the model's predictions for a specific class $a$. A smaller CPIW indicates that the predictions are more tightly clustered, suggesting higher certainty in the model's prediction for that class. Conversely, a larger CPIW suggests greater variability, and thus greater uncertainty.
    \item \textbf{Class-wise Normalized Prediction Variance (CNPV).}\; Calculating the prediction variance is a common method for quantifying uncertainty in medical imaging \cite{Gomes2021Building}. The CNPV is defined as follows:
    \begin{align}\mathrm{CNPV}_a:=|\mathcal{S}_a|^{-1}\sum_{i=1}^{|\mathcal{S}_a|}4(\hat{\mb y}_i^a-\bar{\mb y}^a)^2,
    \end{align}
    where $|\mathcal{S}_a|$ denotes the cardinality of the set $\mathcal{S}_a$, and $\bar{\mb y}^a$ denotes the mean of all values in $\mathcal{S}_a$. CNPV quantifies the variability of the predictions by calculating the normalized variance of the samples in $\mathcal{S}_a$. A lower CNPV value indicates that the predictions are consistent and the model is confident in its decision for that class. Higher CNPV values suggest more uncertainty, as the predictions vary more significantly around the mean.
\end{itemize}

\begin{table}
\centering
\caption{Results of uncertainty measures under covariate shift (Images with Gaussian noise injection $\mathbb{g}=1.00$).}
\label{tab:cpiwcnpv}
\begin{tabular}{c|ccc}
\toprule
                        Class & Evaluated Predictions & CPIW & CNPV \\ \midrule
\multirow{2}{*}{Tuberculosis} & Correct   & $0.4330$   & $0.2520$  \\
                         & Incorrect & $0.8600$   & $0.7196$  \\ \midrule
\multirow{2}{*}{Healthy} & Correct   & $0.9998$   & $0.9112$  \\
                         & Incorrect & $0.9997$   & $0.8940$  \\ 
 \bottomrule
\end{tabular}
\end{table}

\begin{figure}[]
  \centering
  \includegraphics[width=0.7\linewidth]{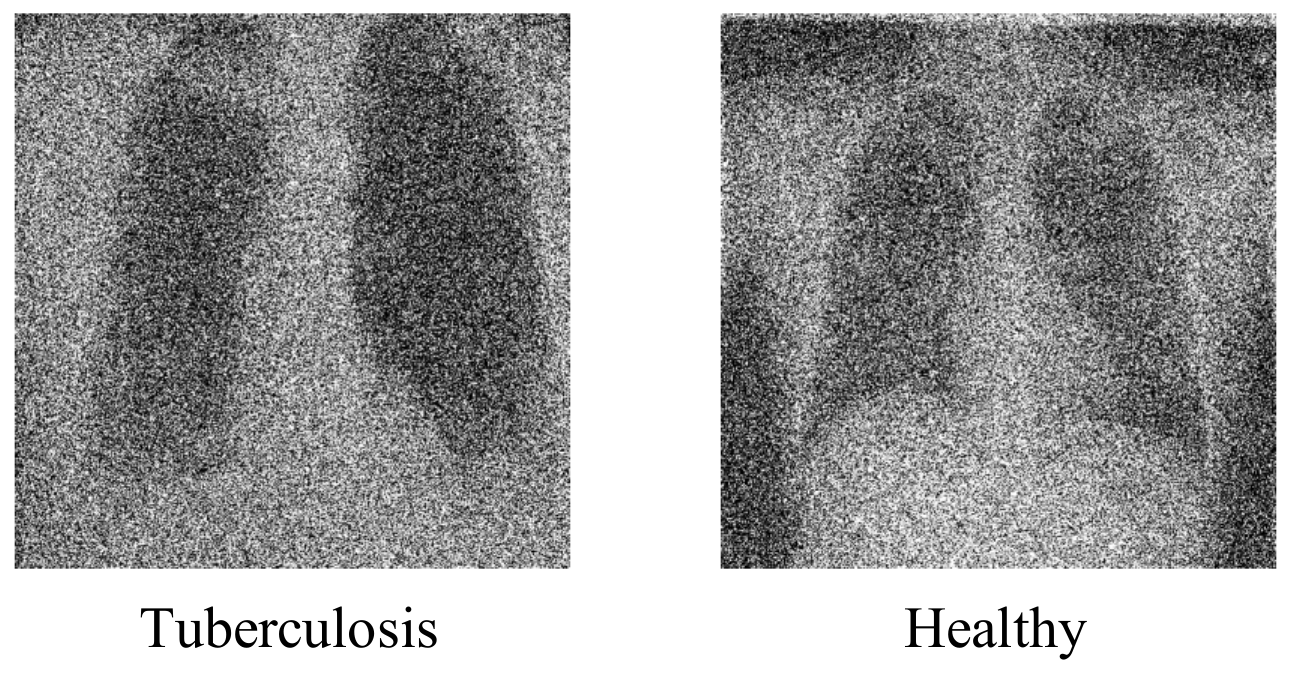}
  \caption{Two images from chest X-ray dataset with Gaussian noise injection ($\mathbb{g}=1.00$): (i) Patient with tuberculosis (left). (ii) Healthy patient (right).}
  \label{fig:tuber_vs_healthy}
\end{figure}

In order to examine the power of the method in challenging contexts, the input images are perturbed {\it significantly} with Gaussian noise ($\mathbb{g}=1.00$), see example images in \cref{fig:tuber_vs_healthy}. This level of  Gaussian noise results in it being challenging to differentiate  healthy images from tuberculosis images. Results shown in~\cref{tab:cpiwcnpv} illustrate the instance-level uncertainty estimates for the LaDiNE's predictions on different classes, specifically focusing on the distinction between correct and incorrect predictions. The first thing to note is that the results reflect how the challenging context results in high uncertainties for the healthy class predictions. LaDiNE appropriately expresses the uncertainty in these challenging cases, informing clinicians to review those uncertain instances more carefully. On the other hand, LaDiNE is more certain when correctly predicting tuberculosis, which demonstrates the model's robustness in detecting true unhealthy cases.

\begin{table*}[h]
\centering
\caption{Comparison in classification accuracy (Acc. \%) and expected calibration error (ECE) with different designs. $M$ indicates the number of sampling times from the distribution. We investigate the effectiveness of inferred latent variable and diffusion model in improving classification accuracy and confidence calibration.}
\label{tab:comp}
\begin{tabular}{cccl|cc|cc}
\toprule
\multirow{2}{*}{\begin{tabular}[c]{@{}c@{}}Design\\ ($K=5$)\end{tabular}} & \multirow{2}{*}{\begin{tabular}[c]{@{}c@{}}With \\ latent\\ variable $z$\end{tabular}} & \multirow{2}{*}{\begin{tabular}[c]{@{}c@{}}With \\ CDM\end{tabular}} & \multirow{2}{*}{\begin{tabular}[c]{@{}l@{}}Functional \\ assumption\end{tabular}} & \multicolumn{2}{c|}{Clean} & \multicolumn{2}{c}{Gaussian noise} \\ \cmidrule{5-8} 
 &  &  &  & Acc. & ECE & \begin{tabular}[c]{@{}c@{}}Acc.\\ (- \% drop compare to Clean)\end{tabular} & ECE \\ \midrule
1 & \cmark & \xmark & Dirac delta (deterministic) (M=1) & 99.71 & 0.3614 & 62.29 (-37.53\%) & 0.3732 \\
2 & \cmark & \xmark & Gaussian (M=20, avg. conf.) & 98.86 & 0.2331 & 61.43 (-37.86\%) & 0.2485 \\
3 & \xmark & \cmark & Any (M=20, avg. conf.) & 99.77 & 0.0031 & 57.42 (-42.44\%) & 0.2273 \\ \midrule
4 & \cmark & \cmark & Any (M=20, avg. conf.) & 99.90 & 0.0030 & 66.33 (-33.60\%) & 0.1960 \\ \bottomrule
\end{tabular}
\end{table*}

\subsection{Ablation Studies}
The effect on classification performance is examined when a variety of other design choices are implemented.

\textbf{Results on Element-wise Ablation Studies.}\; To further justify the design choices made for each component of the framework, the proposed model is tested on four configurations of (i) a clean chest X-ray testing dataset and (ii) a noise injected chest X-ray testing dataset (Gaussian noise with $\mathbb{g}=1.00$):
\begin{itemize}[labelwidth=!, topsep=0pt, leftmargin=0.5cm]
    \item[1.] In this configuration, CDM is removed from LaDiNE and instead a deterministic mapping to the parameters of the final categorical distribution is used. Specifically, the softmaxed latent variable $z$ serves as the predicted class probabilities.
    \item[2.] Instead of learning the distribution with CDM, the CDM is replaced with a Gaussian distribution parameterized by a two-head neural network to estimate the mean and variance of $\mb y$ conditioned on the latent variable $z$. 20 samples are drawn from the Gaussian distribution per ensemble member and the average confidence is estimated.
    \item[3.] To investigate the effectiveness of the inferred latent variable $z$, $z$ is replaced with the output logits from the ViT. 20 samples are drawn from the CDM per ensemble member and the average confidence is estimated.
    \item[4.] The entire LaDiNE is examined, where 20 samples are drawn from the CDM per ensemble member and the average confidence is estimated.
\end{itemize}

As shown in \cref{tab:comp}, the complete version of LaDiNE (design 4) achieves the highest accuracy for all testing sets, the lowest relative accuracy drop under noisy conditions, and the lowest Expected Calibration Error (ECE), providing support and justification for the design elements chosen.

In Design 1 which does not make use of a CDM, the accuracy drops especially under Gaussian noise, and the ECE increases, indicating that the CDM is crucial for robust performance and reliable uncertainty estimation under input image perturbation. Replacing the CDM with a Gaussian distribution (Design 2) also leads to a notable performance degradation, suggesting that the flexibility of the CDM in modeling complex distributions contributes to better accuracy and calibration. On the other hand, Design 2 achieves a lower ECE than Design 1, indicating that encoding predictive confidence (the Gaussian distribution in Design 2) can help mitigate issues with over-confidence.

When the latent variable $z$ is removed (Design 3), the model experiences a severe drop in performance under input image perturbation, emphasizing the importance of $z$ in capturing informative and invariant representations that are critical for generalization and robustness.

Overall, the ablation studies clearly demonstrate that both the CDM and the latent variable $z$ play essential roles in the superior performance of LaDiNE, especially in handling of noisy data from outside the learned distribution.

\textbf{Results on Selection of Mixture Components.}\; Another key design choice in our method is the selection of mixture components, in other words, the TE hierarchy. \cref{tab:accblk} shows the performance of the proposed method with $K=3,4,\dotsc,7$ (we draw 20 samples per ensemble member). When $K=5$, the classification accuracy is the highest for both datasets. When $K$ is smaller, there is insufficient discrimination in the extracted features which therefore results in low classification accuracy. On the other hand, as $K$ increases, the inner structure of feature representations from the different hierarchies becomes too complex and may therefore result in a performance drop.

\begin{table}
\centering
\caption{Comparison in classification accuracy (\%) with different choice of mixture components.}
\label{tab:accblk}
\begin{tabular}{l|c|c}
\toprule
    & Chest X-ray & ISIC  \\ \midrule
$K=3$ & $96.47\pm0.29$       & $87.84\pm0.13$ \\
$K=4$ & $98.63\pm0.60$       & $91.24\pm0.43$ \\
$K=5$ & $\mb{99.90}\pm\mb{0.14}$       & $\mb{94.18}\pm\mb{0.24}$ \\
$K=6$ & $97.59\pm0.25$       & $92.40\pm0.54$ \\
$K=7$ & $98.31\pm0.38$       & $92.14\pm0.41$ \\ \bottomrule
\end{tabular}
\end{table}

Examining the results in~\cref{tab:accblk} and the findings in~\cref{fig:l2}, one can observe a trade-off between informativeness and invariance in selecting early TE blocks' representation. Specifically:
\begin{itemize}[labelwidth=!, topsep=0pt, leftmargin=*]
    \item \textbf{Invariance.}\; Shallow blocks tend to capture more invariant features across different environments, providing a stable representation that is less sensitive to specific variations in the input data. These invariant features are beneficial for maintaining consistency and robustness, especially under covariate shifts. However, these representations may encode less direct information for classifying the input data.
    \item \textbf{Informative.}\; Higher blocks, on the other hand, encode more informative and discriminative features. These features capture more detailed and specific characteristics of the input data, which can enhance classification accuracy. However, this increased specificity can also lead to reduced invariance, and can result poor generalization due to over-parameterization of the mixture.
\end{itemize}

\section{Limitations and Future Works}
One of the limitations of our approach is the increased computational cost associated with diffusion models, which require iterative denoising processes. Although diffusion models are expensive, the denoising process is on $\mathbb{R}^A$ instead of on the whole image space ($A$ is the number of classes). Inference time is still reasonable for many medical imaging tasks. In our experiments, the inference time (on one NVIDIA A100 GPU) per image is approximately $98.23$ ms for LaDiNE (we sequentially sample $K\times M$ times, where $K=5$ and $M=20$), compared to $0.54$ ms for the ViT-B baseline. Future work could explore faster sampling techniques, such as Denoising Diffusion Implicit Models (DDIM) \cite{songdenoising} or consistency models \cite{song2023consistency}, which can reduce the number of required iterations and computational overhead while maintaining comparable performance. In addition to incorporating accelerated sampling techniques, the computation of ensemble members (implemented by $K$ mapping networks and denoising networks) can be parallelized to further reduce the latency and improve efficiency. By distributing the ensemble's workload across multiple processing units (e.g., multiple graphical processing units), inference times can be significantly reduced, making the approach more efficient and practical for various applications.

Another limitation is the slight variability in performance depending on the neural network initialization compared to other methods. This sensitivity can lead to inconsistencies in model outputs. Future research could integrate Bayesian deep learning techniques, which explicitly model uncertainty in the network parameters. Approaches such as Bayesian neural networks (BNNs) or approximate Bayesian inference~\cite{gal2016dropout} can provide more reliable uncertainty estimates and help stabilize performance across different initializations.

\section{Conclusion}
In this work, we present a novel ensemble learning approach, LaDiNE, designed to improve the robustness and reliability of medical image classification under covariate shifts. By learning invariant features and modeling the predictive distribution with a functional-form-free mixture, the proposed approach effectively addresses the challenges of image perturbations and adversarial attacks on the inputs, and achieving calibrated confidence levels in its predictions. Extensive experiments on benchmark datasets demonstrate the superiority of LaDiNE in achieving high classification accuracy and well-calibrated prediction confidence under various challenging conditions. This work underscores the importance of robust and reliable models in clinical decision-making, providing a pathway for future advancements in trustworthy artificial intelligence for medical image analysis.

{\small
\bibliographystyle{IEEEtran}
\bibliography{egbib}

\begin{thebibliography}{10}
\providecommand{\url}[1]{#1}
\csname url@samestyle\endcsname
\providecommand{\newblock}{\relax}
\providecommand{\bibinfo}[2]{#2}
\providecommand{\BIBentrySTDinterwordspacing}{\spaceskip=0pt\relax}
\providecommand{\BIBentryALTinterwordstretchfactor}{4}
\providecommand{\BIBentryALTinterwordspacing}{\spaceskip=\fontdimen2\font plus
\BIBentryALTinterwordstretchfactor\fontdimen3\font minus \fontdimen4\font\relax}
\providecommand{\BIBforeignlanguage}[2]{{%
\expandafter\ifx\csname l@#1\endcsname\relax
\typeout{** WARNING: IEEEtran.bst: No hyphenation pattern has been}%
\typeout{** loaded for the language `#1'. Using the pattern for}%
\typeout{** the default language instead.}%
\else
\language=\csname l@#1\endcsname
\fi
#2}}
\providecommand{\BIBdecl}{\relax}
\BIBdecl

\bibitem{esteva2017dermatologist}
A.~Esteva, B.~Kuprel, R.~A. Novoa, J.~Ko, S.~M. Swetter, H.~M. Blau, and S.~Thrun, ``Dermatologist-level classification of skin cancer with deep neural networks,'' \emph{Nature}, vol. 542, no. 7639, pp. 115--118, 2017.

\bibitem{gulshan2016development}
V.~Gulshan, L.~Peng, M.~Coram, M.~C. Stumpe, D.~Wu, A.~Narayanaswamy, S.~Venugopalan, K.~Widner, T.~Madams, J.~Cuadros \emph{et~al.}, ``Development and validation of a deep learning algorithm for detection of diabetic retinopathy in retinal fundus photographs,'' \emph{Journal of the American Medical Association}, vol. 316, no.~22, pp. 2402--2410, 2016.

\bibitem{ardila2019end}
D.~Ardila, A.~P. Kiraly, S.~Bharadwaj, B.~Choi, J.~J. Reicher, L.~Peng, D.~Tse, M.~Etemadi, W.~Ye, G.~Corrado \emph{et~al.}, ``End-to-end lung cancer screening with three-dimensional deep learning on low-dose chest computed tomography,'' \emph{Nature Medicine}, vol.~25, no.~6, pp. 954--961, 2019.

\bibitem{bejnordi2017diagnostic}
B.~E. Bejnordi, M.~Veta, P.~J. Van~Diest, B.~Van~Ginneken, N.~Karssemeijer, G.~Litjens, J.~A. Van Der~Laak, M.~Hermsen, Q.~F. Manson, M.~Balkenhol \emph{et~al.}, ``Diagnostic assessment of deep learning algorithms for detection of lymph node metastases in women with breast cancer,'' \emph{Journal of the American Medical Association}, vol. 318, no.~22, pp. 2199--2210, 2017.

\bibitem{de2018clinically}
J.~De~Fauw, J.~R. Ledsam, B.~Romera-Paredes, S.~Nikolov, N.~Tomasev, S.~Blackwell, H.~Askham, X.~Glorot, B.~O’Donoghue, D.~Visentin \emph{et~al.}, ``Clinically applicable deep learning for diagnosis and referral in retinal disease,'' \emph{Nature Medicine}, vol.~24, no.~9, pp. 1342--1350, 2018.

\bibitem{litjens2017survey}
G.~Litjens, T.~Kooi, B.~E. Bejnordi, A.~A.~A. Setio, F.~Ciompi, M.~Ghafoorian, J.~A. Van Der~Laak, B.~Van~Ginneken, and C.~I. S{\'a}nchez, ``A survey on deep learning in medical image analysis,'' \emph{Medical Image Analysis}, vol.~42, pp. 60--88, 2017.

\bibitem{qiu2016initial}
Y.~Qiu, Y.~Wang, S.~Yan, M.~Tan, S.~Cheng, H.~Liu, and B.~Zheng, ``An initial investigation on developing a new method to predict short-term breast cancer risk based on deep learning technology,'' in \emph{Medical Imaging 2016: Computer-Aided Diagnosis}, vol. 9785.\hskip 1em plus 0.5em minus 0.4em\relax SPIE, 2016, pp. 517--522.

\bibitem{tajbakhsh2016convolutional}
N.~Tajbakhsh, J.~Y. Shin, S.~R. Gurudu, R.~T. Hurst, C.~B. Kendall, M.~B. Gotway, and J.~Liang, ``Convolutional neural networks for medical image analysis: Full training or fine tuning?'' \emph{IEEE Transactions on Medical Imaging}, vol.~35, no.~5, pp. 1299--1312, 2016.

\bibitem{milletari2016v}
F.~Milletari, N.~Navab, and S.-A. Ahmadi, ``V-net: Fully convolutional neural networks for volumetric medical image segmentation,'' in \emph{International Conference on 3D Vision}.\hskip 1em plus 0.5em minus 0.4em\relax IEEE, 2016, pp. 565--571.

\bibitem{hendrycks2018benchmarking}
D.~Hendrycks and T.~Dietterich, ``Benchmarking neural network robustness to common corruptions and perturbations,'' in \emph{International Conference on Learning Representations}, 2018.

\bibitem{szegedy2014intriguingpropertiesneuralnetworks}
C.~Szegedy, W.~Zaremba, I.~Sutskever, J.~Bruna, D.~Erhan, I.~Goodfellow, and R.~Fergus, ``Intriguing properties of neural networks,'' in \emph{International Conference on Learning Representations}, 2014.

\bibitem{navarro2021evaluating}
F.~Navarro, C.~Watanabe, S.~Shit, A.~Sekuboyina, J.~C. Peeken, S.~E. Combs, and B.~H. Menze, ``Evaluating the robustness of self-supervised learning in medical imaging,'' \emph{arXiv preprint arXiv:2105.06986}, 2021.

\bibitem{roglin2022improving}
J.~R{\"o}glin, K.~Ziegeler, J.~Kube, F.~K{\"o}nig, K.-G. Hermann, and S.~Ortmann, ``Improving classification results on a small medical dataset using a gan; an outlook for dealing with rare disease datasets,'' \emph{Frontiers in Computer Science}, vol.~4, p. 858874, 2022.

\bibitem{takase2021self}
T.~Takase, R.~Karakida, and H.~Asoh, ``Self-paced data augmentation for training neural networks,'' \emph{Neurocomputing}, vol. 442, pp. 296--306, 2021.

\bibitem{matta2024systematic}
S.~Matta, M.~Lamard, P.~Zhang, A.~Le~Guilcher, L.~Borderie, B.~Cochener, and G.~Quellec, ``A systematic review of generalization research in medical image classification,'' \emph{Computers in biology and medicine}, vol. 183, p. 109256, 2024.

\bibitem{breiman1996bagging}
L.~Breiman, ``Bagging predictors,'' \emph{Machine Learning}, vol.~24, pp. 123--140, 1996.

\bibitem{Mienye2022A}
I.~D. Mienye and Y.~Sun, ``A survey of ensemble learning: Concepts, algorithms, applications, and prospects,'' \emph{IEEE Access}, vol.~10, pp. 99\,129--99\,149, 2022.

\bibitem{lakshminarayanan2017simple}
B.~Lakshminarayanan, A.~Pritzel, and C.~Blundell, ``Simple and scalable predictive uncertainty estimation using deep ensembles,'' in \emph{Advances in Neural Information Processing Systems}, vol.~30, 2017.

\bibitem{yang2021two}
Y.~Yang, Y.~Hu, X.~Zhang, and S.~Wang, ``Two-stage selective ensemble of cnn via deep tree training for medical image classification,'' \emph{IEEE Transactions on Cybernetics}, vol.~52, no.~9, pp. 9194--9207, 2021.

\bibitem{Pacheco2020Learning}
A.~G.~C. Pacheco, T.~Trappenberg, and R.~Krohling, ``Learning dynamic weights for an ensemble of deep models applied to medical imaging classification,'' \emph{2020 International Joint Conference on Neural Networks (IJCNN)}, pp. 1--8, 2020.

\bibitem{shamshad2023transformers}
F.~Shamshad, S.~Khan, S.~W. Zamir, M.~H. Khan, M.~Hayat, F.~S. Khan, and H.~Fu, ``Transformers in medical imaging: A survey,'' \emph{Medical Image Analysis}, p. 102802, 2023.

\bibitem{kazerouni2023diffusion}
A.~Kazerouni, E.~K. Aghdam, M.~Heidari, R.~Azad, M.~Fayyaz, I.~Hacihaliloglu, and D.~Merhof, ``Diffusion models in medical imaging: A comprehensive survey,'' \emph{Medical Image Analysis}, p. 102846, 2023.

\bibitem{rahman2020reliable}
T.~Rahman, A.~Khandakar, M.~A. Kadir, K.~R. Islam, K.~F. Islam, R.~Mazhar, T.~Hamid, M.~T. Islam, S.~Kashem, Z.~B. Mahbub \emph{et~al.}, ``Reliable tuberculosis detection using chest x-ray with deep learning, segmentation and visualization,'' \emph{IEEE Access}, vol.~8, pp. 191\,586--191\,601, 2020.

\bibitem{rotemberg2021patient}
V.~Rotemberg, N.~Kurtansky, B.~Betz-Stablein, L.~Caffery, E.~Chousakos, N.~Codella, M.~Combalia, S.~Dusza, P.~Guitera, D.~Gutman \emph{et~al.}, ``A patient-centric dataset of images and metadata for identifying melanomas using clinical context,'' \emph{Scientific Data}, vol.~8, no.~1, p.~34, 2021.

\bibitem{peiris2022robust}
H.~Peiris, M.~Hayat, Z.~Chen, G.~Egan, and M.~Harandi, ``A robust volumetric transformer for accurate 3d tumor segmentation,'' in \emph{International conference on medical image computing and computer-assisted intervention}.\hskip 1em plus 0.5em minus 0.4em\relax Springer, 2022, pp. 162--172.

\bibitem{chen2022gashis}
H.~Chen, C.~Li, G.~Wang, X.~Li, M.~M. Rahaman, H.~Sun, W.~Hu, Y.~Li, W.~Liu, C.~Sun \emph{et~al.}, ``Gashis-transformer: A multi-scale visual transformer approach for gastric histopathological image detection,'' \emph{Pattern Recognition}, vol. 130, p. 108827, 2022.

\bibitem{wang2021dudotrans}
C.~Wang, K.~Shang, H.~Zhang, Q.~Li, Y.~Hui, and S.~K. Zhou, ``Dudotrans: dual-domain transformer provides more attention for sinogram restoration in sparse-view ct reconstruction,'' \emph{arXiv preprint arXiv:2111.10790}, 2021.

\bibitem{almalik2022self}
F.~Almalik, M.~Yaqub, and K.~Nandakumar, ``Self-ensembling vision transformer (sevit) for robust medical image classification,'' in \emph{Medical Image Computing and Computer Assisted Intervention--MICCAI 2022: 25th International Conference, Singapore, September 18--22, 2022, Proceedings, Part III}.\hskip 1em plus 0.5em minus 0.4em\relax Springer, 2022, pp. 376--386.

\bibitem{dhariwal2021diffusion}
P.~Dhariwal and A.~Nichol, ``Diffusion models beat gans on image synthesis,'' \emph{Advances in neural information processing systems}, vol.~34, pp. 8780--8794, 2021.

\bibitem{pinaya2022brain}
W.~H. Pinaya, P.-D. Tudosiu, J.~Dafflon, P.~F. Da~Costa, V.~Fernandez, P.~Nachev, S.~Ourselin, and M.~J. Cardoso, ``Brain imaging generation with latent diffusion models,'' in \emph{MICCAI Workshop on Deep Generative Models}.\hskip 1em plus 0.5em minus 0.4em\relax Springer, 2022, pp. 117--126.

\bibitem{moghadam2023morphology}
P.~A. Moghadam, S.~Van~Dalen, K.~C. Martin, J.~Lennerz, S.~Yip, H.~Farahani, and A.~Bashashati, ``A morphology focused diffusion probabilistic model for synthesis of histopathology images,'' in \emph{Proceedings of the IEEE/CVF winter conference on applications of computer vision}, 2023, pp. 2000--2009.

\bibitem{yoon2023sadm}
J.~S. Yoon, C.~Zhang, H.-I. Suk, J.~Guo, and X.~Li, ``Sadm: Sequence-aware diffusion model for longitudinal medical image generation,'' in \emph{International Conference on Information Processing in Medical Imaging}.\hskip 1em plus 0.5em minus 0.4em\relax Springer, 2023, pp. 388--400.

\bibitem{song2021solving}
Y.~Song, L.~Shen, L.~Xing, and S.~Ermon, ``Solving inverse problems in medical imaging with score-based generative models,'' in \emph{International Conference on Learning Representations}, 2021.

\bibitem{xie2022measurement}
Y.~Xie and Q.~Li, ``Measurement-conditioned denoising diffusion probabilistic model for under-sampled medical image reconstruction,'' in \emph{International Conference on Medical Image Computing and Computer-Assisted Intervention}.\hskip 1em plus 0.5em minus 0.4em\relax Springer, 2022, pp. 655--664.

\bibitem{wu2024medsegdiff}
J.~Wu, R.~Fu, H.~Fang, Y.~Zhang, Y.~Yang, H.~Xiong, H.~Liu, and Y.~Xu, ``Medsegdiff: Medical image segmentation with diffusion probabilistic model,'' in \emph{Medical Imaging with Deep Learning}.\hskip 1em plus 0.5em minus 0.4em\relax PMLR, 2024, pp. 1623--1639.

\bibitem{wolleb2022diffusion}
J.~Wolleb, R.~Sandk{\"u}hler, F.~Bieder, P.~Valmaggia, and P.~C. Cattin, ``Diffusion models for implicit image segmentation ensembles,'' in \emph{International Conference on Medical Imaging with Deep Learning}.\hskip 1em plus 0.5em minus 0.4em\relax PMLR, 2022, pp. 1336--1348.

\bibitem{li2023zero}
Y.~Li, H.-C. Shao, X.~Liang, L.~Chen, R.~Li, S.~Jiang, J.~Wang, and Y.~Zhang, ``Zero-shot medical image translation via frequency-guided diffusion models,'' \emph{IEEE Transactions on Medical Imaging}, 2023.

\bibitem{kim2022diffusion}
B.~Kim, Y.~Oh, and J.~C. Ye, ``Diffusion adversarial representation learning for self-supervised vessel segmentation,'' in \emph{International Conference on Learning Representations}, 2022.

\bibitem{clark2024text}
K.~Clark and P.~Jaini, ``Text-to-image diffusion models are zero shot classifiers,'' \emph{Advances in Neural Information Processing Systems}, vol.~36, 2024.

\bibitem{han2022card}
X.~Han, H.~Zheng, and M.~Zhou, ``Card: Classification and regression diffusion models,'' in \emph{Advances in Neural Information Processing Systems}, vol.~35, 2022, pp. 18\,100--18\,115.

\bibitem{chendiffusion}
H.~Chen, Y.~Dong, S.~Shao, Z.~Hao, X.~Yang, H.~Su, and J.~Zhu, ``Diffusion models are certifiably robust classifiers,'' in \emph{The Thirty-eighth Annual Conference on Neural Information Processing Systems}, 2024.

\bibitem{dosovitskiyimage}
A.~Dosovitskiy, L.~Beyer, A.~Kolesnikov, D.~Weissenborn, X.~Zhai, T.~Unterthiner, M.~Dehghani, M.~Minderer, G.~Heigold, S.~Gelly \emph{et~al.}, ``An image is worth 16x16 words: Transformers for image recognition at scale,'' in \emph{International Conference on Learning Representations}, 2021.

\bibitem{walmer2023teaching}
M.~Walmer, S.~Suri, K.~Gupta, and A.~Shrivastava, ``Teaching matters: Investigating the role of supervision in vision transformers,'' in \emph{Proceedings of the IEEE/CVF Conference on Computer Vision and Pattern Recognition}, 2023, pp. 7486--7496.

\bibitem{ho2020denoising}
J.~Ho, A.~Jain, and P.~Abbeel, ``Denoising diffusion probabilistic models,'' in \emph{Advances in Neural Information Processing Systems}, vol.~33, 2020, pp. 6840--6851.

\bibitem{luo2022understanding}
C.~Luo, ``Understanding diffusion models: A unified perspective,'' \emph{arXiv preprint arXiv:2208.11970}, 2022.

\bibitem{karras2022elucidating}
T.~Karras, M.~Aittala, T.~Aila, and S.~Laine, ``Elucidating the design space of diffusion-based generative models,'' \emph{Advances in Neural Information Processing Systems}, vol.~35, pp. 26\,565--26\,577, 2022.

\bibitem{brier1950verification}
G.~W. Brier, ``Verification of forecasts expressed in terms of probability,'' \emph{Monthly Weather Review}, vol.~78, no.~1, pp. 1--3, 1950.

\bibitem{he2016deep}
K.~He, X.~Zhang, S.~Ren, and J.~Sun, ``Deep residual learning for image recognition,'' in \emph{Proceedings of the IEEE/CVF Conference on Computer Vision and Pattern Recognition}, 2016, pp. 770--778.

\bibitem{manzari2023medvit}
O.~N. Manzari, H.~Ahmadabadi, H.~Kashiani, S.~B. Shokouhi, and A.~Ayatollahi, ``Medvit: a robust vision transformer for generalized medical image classification,'' \emph{Computers in Biology and Medicine}, vol. 157, p. 106791, 2023.

\bibitem{musaev2023icnn}
J.~Musaev, A.~Anorboev, Y.-S. Seo, N.~T. Nguyen, and D.~Hwang, ``Icnn-ensemble: An improved convolutional neural network ensemble model for medical image classification,'' \emph{IEEE Access}, 2023.

\bibitem{nelder1965simplex}
J.~A. Nelder and R.~Mead, ``A simplex method for function minimization,'' \emph{The Computer Journal}, vol.~7, no.~4, pp. 308--313, 1965.

\bibitem{tan2021efficientnetv2}
M.~Tan and Q.~Le, ``Efficientnetv2: Smaller models and faster training,'' in \emph{International Conference on Machine Learning}.\hskip 1em plus 0.5em minus 0.4em\relax PMLR, 2021, pp. 10\,096--10\,106.

\bibitem{touvron2021training}
H.~Touvron, M.~Cord, M.~Douze, F.~Massa, A.~Sablayrolles, and H.~J{\'e}gou, ``Training data-efficient image transformers \& distillation through attention,'' in \emph{International Conference on Machine Learning}.\hskip 1em plus 0.5em minus 0.4em\relax PMLR, 2021, pp. 10\,347--10\,357.

\bibitem{liu2021swin}
Z.~Liu, Y.~Lin, Y.~Cao, H.~Hu, Y.~Wei, Z.~Zhang, S.~Lin, and B.~Guo, ``Swin transformer: Hierarchical vision transformer using shifted windows,'' in \emph{Proceedings of the IEEE/CVF International Conference on Computer Vision}, 2021, pp. 10\,012--10\,022.

\bibitem{d2021convit}
S.~d’Ascoli, H.~Touvron, M.~L. Leavitt, A.~S. Morcos, G.~Biroli, and L.~Sagun, ``Convit: Improving vision transformers with soft convolutional inductive biases,'' in \emph{International Conference on Machine Learning}.\hskip 1em plus 0.5em minus 0.4em\relax PMLR, 2021, pp. 2286--2296.

\bibitem{finlayson2019adversarial}
S.~G. Finlayson, J.~D. Bowers, J.~Ito, J.~L. Zittrain, A.~L. Beam, and I.~S. Kohane, ``Adversarial attacks on medical machine learning,'' \emph{Science}, vol. 363, no. 6433, pp. 1287--1289, 2019.

\bibitem{kaviani2022adversarial}
S.~Kaviani, K.~J. Han, and I.~Sohn, ``Adversarial attacks and defenses on ai in medical imaging informatics: A survey,'' \emph{Expert Systems with Applications}, vol. 198, p. 116815, 2022.

\bibitem{bortsova2021adversarial}
G.~Bortsova, C.~Gonz{\'a}lez-Gonzalo, S.~C. Wetstein, F.~Dubost, I.~Katramados, L.~Hogeweg, B.~Liefers, B.~van Ginneken, J.~P. Pluim, M.~Veta \emph{et~al.}, ``Adversarial attack vulnerability of medical image analysis systems: Unexplored factors,'' \emph{Medical Image Analysis}, vol.~73, p. 102141, 2021.

\bibitem{goodfellow2014explaining}
I.~J. Goodfellow, J.~Shlens, and C.~Szegedy, ``Explaining and harnessing adversarial examples,'' in \emph{International Conference on Learning Representations}, 2015.

\bibitem{madry2018towards}
A.~Madry, A.~Makelov, L.~Schmidt, D.~Tsipras, and A.~Vladu, ``Towards deep learning models resistant to adversarial attacks,'' in \emph{International Conference on Learning Representations}, 2018.

\bibitem{croce2020reliable}
F.~Croce and M.~Hein, ``Reliable evaluation of adversarial robustness with an ensemble of diverse parameter-free attacks,'' in \emph{International Conference on Machine Learning}.\hskip 1em plus 0.5em minus 0.4em\relax PMLR, 2020, pp. 2206--2216.

\bibitem{naeini2015obtaining}
M.~P. Naeini, G.~Cooper, and M.~Hauskrecht, ``Obtaining well calibrated probabilities using bayesian binning,'' in \emph{Proceedings of the AAAI Conference on Artificial Intelligence}, vol.~29, no.~1, 2015.

\bibitem{guo2017calibration}
C.~Guo, G.~Pleiss, Y.~Sun, and K.~Q. Weinberger, ``On calibration of modern neural networks,'' in \emph{International Conference on Machine Learning}.\hskip 1em plus 0.5em minus 0.4em\relax PMLR, 2017, pp. 1321--1330.

\bibitem{jungo2020analyzing}
A.~Jungo, F.~Balsiger, and M.~Reyes, ``Analyzing the quality and challenges of uncertainty estimations for brain tumor segmentation,'' \emph{Frontiers in Neuroscience}, vol.~14, p. 282, 2020.

\bibitem{shui2023mitigating}
C.~Shui, J.~Szeto, R.~Mehta, D.~L. Arnold, and T.~Arbel, ``Mitigating calibration bias without fixed attribute grouping for improved fairness in medical imaging analysis,'' in \emph{International Conference on Medical Image Computing and Computer-Assisted Intervention}.\hskip 1em plus 0.5em minus 0.4em\relax Springer, 2023, pp. 189--198.

\bibitem{mehta2022qu}
R.~Mehta, A.~Filos, U.~Baid, C.~Sako, R.~McKinley, M.~Rebsamen, K.~D{\"a}twyler, R.~Meier, P.~Radojewski, G.~K. Murugesan \emph{et~al.}, ``Qu-brats: Miccai brats 2020 challenge on quantifying uncertainty in brain tumor segmentation-analysis of ranking scores and benchmarking results,'' \emph{The journal of machine learning for biomedical imaging}, vol. 2022, 2022.

\bibitem{Gomes2021Building}
J.~M. Gomes, J.~Kong, T.~Kurç, A.~C. Melo, R.~Ferreira, J.~Saltz, and G.~Teodoro, ``Building robust pathology image analyses with uncertainty quantification,'' \emph{Computer Methods and Programs in Biomedicine}, vol. 208, p. 106291, 2021.

\bibitem{songdenoising}
J.~Song, C.~Meng, and S.~Ermon, ``Denoising diffusion implicit models,'' in \emph{International Conference on Learning Representations}, 2021.

\bibitem{song2023consistency}
Y.~Song, P.~Dhariwal, M.~Chen, and I.~Sutskever, ``Consistency models,'' in \emph{International Conference on Machine Learning}.\hskip 1em plus 0.5em minus 0.4em\relax PMLR, 2023, pp. 32\,211--32\,252.

\bibitem{gal2016dropout}
Y.~Gal and Z.~Ghahramani, ``Dropout as a bayesian approximation: Representing model uncertainty in deep learning,'' in \emph{International Conference on Machine Learning}.\hskip 1em plus 0.5em minus 0.4em\relax PMLR, 2016, pp. 1050--1059.

\end{thebibliography}
}

\end{document}